\documentclass{article}

\usepackage{arxiv}

\usepackage[utf8]{inputenc} 
\usepackage[T1]{fontenc}    
\usepackage{hyperref}       
\usepackage{url}            
\usepackage{booktabs}       
\usepackage{amsmath}        
\usepackage{amssymb}        
\usepackage{amsfonts}       
\usepackage{amsthm}         
\usepackage{nicefrac}       
\usepackage{microtype}      
\usepackage{xcolor}         
\usepackage{graphicx}       
\usepackage{multirow}       
\usepackage{natbib}         
\usepackage{doi}

\newtheorem{lemma}{Lemma}
\newtheorem{theorem}{Theorem}
\newtheorem{corollary}{Corollary}
\theoremstyle{remark}

\newcommand{\R}{\mathbb{R}}
\newcommand{\E}{\mathbb{E}}

\DeclareMathOperator{\Unif}{Unif}

\title{Constant-Target Energy Matching:\\ A Unified Framework for Continuous and Discrete Density Estimation}

\author{%
  Zhijun Zeng\thanks{Equal contribution.} \\
  Department of Mathematical Sciences\\
  Tsinghua University\\
  Beijing 100084, China
  \And
  Yixuan Jiang\footnotemark[1] \\
  Qiuzhen College\\
  Tsinghua University\\
  Beijing 100084, China
  \And
  Pipi Hu \\
  Yanqi Lake Beijing Institute of\\
  Mathematical Sciences and Applications\\
  Beijing 101408, China
  \And
  Zuoqiang Shi\thanks{Corresponding author.} \\
  Yau Mathematical Sciences Center\\
  Tsinghua University, Beijing 100084, China\\
  and Yanqi Lake Beijing Institute of\\
  Mathematical Sciences and Applications\\
  Beijing 101408, China\\
  \texttt{zqshi@tsinghua.edu.cn}
}

\renewcommand{\shorttitle}{Constant-Target Energy Matching}

\hypersetup{
pdftitle={Constant-Target Energy Matching: A Unified Framework for Continuous and Discrete Density Estimation},
pdfauthor={Zhijun Zeng, Yixuan Jiang, Pipi Hu, Zuoqiang Shi},
pdfsubject={Equal contribution by Zhijun Zeng and Yixuan Jiang. Corresponding author: Zuoqiang Shi (zqshi@tsinghua.edu.cn).},
pdfkeywords={density estimation, energy-based models, score matching, discrete diffusion, generative models},
}

\begin{document}

\maketitle

\begin{abstract}
Density estimation is a central primitive in probabilistic modeling, yet
continuous, discrete, and mixed-variable domains are often treated by separate
objectives, limiting the ability to exploit a common statistical structure
across data types.
Continuous score-based methods rely on log-density gradients, while discrete
extensions typically use concrete score whose unbounded targets become
unstable near low-probability states. We introduce \emph{Constant-Target Energy Matching}
(CTEM), a unified energy-based framework for density estimation on general
state spaces. CTEM replaces ordinary density-ratio regression with a bounded
energy-difference transform and derives from it a sample-only training objective
with the constant target $1$. The learned scalar potential recovers
$\log\rho$  without partition-function estimation or
explicit unbounded ratio regression. Across continuous, discrete, and mixed-variable benchmarks, CTEM substantially improves density estimation over competitive
baselines and yields higher-quality samples under standard sampling
procedures.
\end{abstract}

\section{Introduction}
\label{sec:intro}

Density estimation is a fundamental primitive in probabilistic modeling,
sampling, inference, and uncertainty quantification. Classical kernel
estimators~\citep{rosenblatt1956remarks,silverman1986density},
energy-based models~\citep{lecun2006tutorial,du2019implicit}, normalizing
flows~\citep{rezende2015variational,dinh2017density,papamakarios2017masked,
kingma2018glow,grathwohl2019ffjord}, and score-based generative
models~\citep{hyvarinen2005estimation,vincent2011connection,
song2019sliced,song2019generative,ho2020denoising,song2021score,
karras2022elucidating} provide complementary ways to represent and learn
probability distributions. Despite this progress, most modern objectives remain
closely tied to the type of state space on which the distribution is defined:
continuous methods exploit differential structure through scores, while
discrete methods typically rely on finite differences, probability ratios, or
denoising surrogates over neighboring states.

This separation is especially visible in score-based modeling. In continuous
spaces, score matching avoids partition-function estimation by learning
$\nabla\log\rho$ instead of the normalized density itself
~\citep{hyvarinen2005estimation,vincent2011connection,song2019sliced}.
This idea underlies modern diffusion and score-based generative models
~\citep{song2019generative,ho2020denoising,song2021score}. However, the score
$\nabla\log\rho$ is not defined on general discrete spaces. Recent discrete
score and diffusion methods therefore replace gradients by finite-difference or
ratio-based quantities over structured neighborhoods
~\citep{hyvarinen2007some,austin2021structured,campbell2022continuous,
meng2022concrete,lou2024discrete,sahoo2024simple,shi2024simplified,
gat2024discrete,sahoo2025diffusion}. These approaches have enabled substantial
progress, but their regression targets often involve ordinary density ratios
such as $\rho(\tilde z)/\rho(z)$, or conditional surrogates of such ratios.

Ordinary ratios are statistically natural but numerically brittle: they are
unbounded and become poorly conditioned near low-probability states. This issue
is particularly acute on sparse discrete supports and mixed spaces, where many
local comparisons involve states with highly unequal probability mass. Existing
discrete generative models address this instability through score-entropy
objectives~\citep{lou2024discrete}, absorbing-state or masked parameterizations
~\citep{sahoo2024simple,shi2024simplified,ou2024absorbing}, flow-based
formulations~\citep{gat2024discrete}, or noise-process designs that connect
discrete and Gaussian diffusion~\citep{sahoo2025diffusion}. While effective in
their intended settings, these remedies are usually coupled to a particular
state-space geometry, corruption process, or reverse-process parameterization.
They therefore do not directly provide a single density-estimation objective
that applies unchanged across continuous, discrete, and mixed-variable domains.

We introduce \emph{Constant-Target Energy Matching} (CTEM), a unified
energy-based framework for density estimation on general state spaces. CTEM
learns a scalar potential $f_\theta\approx \log\rho+C$ by replacing the
ordinary density-ratio target with the bounded modified ratio
\[
    \frac{\rho(z)-\rho(\tilde z)}
         {\rho(z)+\rho(\tilde z)}
    =
    \tanh\!\left(
        \frac{\log\rho(z)-\log\rho(\tilde z)}{2}
    \right).
\]
Starting from a density-weighted matching objective over pairs of states, we
show that a symmetry argument removes the unknown density terms and yields a
sample-only loss with the constant regression target $1$. The same objective
applies to continuous variables through local directional or Gaussian
comparisons, to discrete variables through graph-based neighborhood
comparisons, and to mixed product spaces by combining the two. At the population
level, the learned potential recovers $\log\rho$ up to an additive constant on
each connected component induced by the comparison structure.

Empirically, CTEM performs well across continuous, discrete, and mixed-variable
settings. On continuous density-estimation benchmarks, CTEM improves density
recovery over kernel estimators and normalizing-flow baselines
~\citep{silverman1986density,bowman1984,dinh2017density,durkan2019neural,
epstein2025score}.  On sparse quantized distributions, CTEM reduces total
variation and KL divergence relative to concrete score matching and
SEDD~\citep{meng2022concrete,lou2024discrete}. On binary MNIST, it produces
recognizable samples using a noise-conditional scalar energy and
Metropolis--Hastings sampling. On mixed continuous--discrete product spaces, it
recovers a globally aligned energy over both Euclidean and categorical
coordinates.

\paragraph{Contributions.}
\begin{enumerate}
\item We propose \emph{Constant-Target Energy Matching} (CTEM), a unified
density-estimation framework for continuous, discrete, and mixed state spaces.
CTEM uses a single scalar-potential parameterization and a common training
principle across heterogeneous variables, providing a direct route to
multimodal probabilistic modeling beyond purely continuous or purely discrete
domains.

\item We derive a contrastive density-estimation loss from a modified
density-ratio identity. The resulting objective is sample-based, applies to
both continuous and discrete variables, and avoids explicit regression to
unbounded density ratios that become unstable near low-density states. Unlike
small-noise score-matching objectives, the population target is not tied to an
infinitesimal perturbation limit, so the comparison scale can be chosen for
training stability and coverage.

\item We validate CTEM across continuous, discrete, and mixed-variable density
estimation tasks. CTEM improves density estimation over competitive kernel,
flow, and discrete score-matching baselines; the induced score  yields high-quality samples under standard sampling procedures; and the
same objective remains effective on mixed continuous--discrete product spaces.
\end{enumerate}

\section{Preliminaries}
\label{sec:preliminaries}
Let
$(\mathcal{Z},\nu)$ be a measurable space with a $\sigma$-finite reference
measure, covering continuous variables, discrete variables, and their products
by choosing $\nu$ as the corresponding Lebesgue, counting, or product reference
measure. Let $P$ be the data distribution and assume that $P\ll\nu$ with
strictly positive density
\begin{equation}
    \rho(z)=\frac{dP}{d\nu}(z).
    \label{eq:rho-def}
\end{equation}
We learn a scalar energy $f_\theta:\mathcal{Z}\to\R$ such that
\begin{equation}
    f_\theta(z)=\log\rho(z)+C,
    \label{eq:energy-target}
\end{equation}
where the additive constant is unidentifiable. In discrete spaces, unless
otherwise specified, we take $\nu$ to be the counting measure, so that
$\rho(z_i)=P(z_i)$ and \eqref{eq:energy-target} recovers the log-probability
mass up to an additive constant.

\section{Method}
\label{sec:method}

We introduce \emph{Constant-Target Energy Matching} (CTEM), an
energy-based framework for density estimation on general state spaces.  Existing ratio-based methods, including ratio matching and concrete score
matching, often target ordinary density ratios such as
$\rho(\tilde z)/\rho(z)$. These ratios encode relative probabilities, but they
are unbounded and can be numerically unstable when the denominator is small.
CTEM instead uses the bounded \emph{modified density ratio}
\begin{equation}
    r_\rho(z,\tilde z)
    :=
    \frac{\rho(z)-\rho(\tilde z)}
         {\rho(z)+\rho(\tilde z)},
    \label{eq:modified-density-ratio}
\end{equation}
which is linked to log-density differences by the following identity.

\begin{lemma}[Bounded modified-ratio identity]
\label{lem:identity}
For any $z,\tilde z\in\mathcal{Z}$ with
$\rho(z),\rho(\tilde z)>0$,
\begin{equation}
    \tanh\!\left(
        \frac{\log\rho(z)-\log\rho(\tilde z)}{2}
    \right)
    =
    \frac{\rho(z)-\rho(\tilde z)}
         {\rho(z)+\rho(\tilde z)} .
    \label{eq:bounded-identity}
\end{equation}
\end{lemma}

Lemma~\ref{lem:identity} suggests learning $f_\theta$ by matching the
model-side bounded difference
\[
    \tanh\!\left(
        \frac{f_\theta(z)-f_\theta(\tilde z)}{2}
    \right)
\]
to $r_\rho(z,\tilde z)$. The challenge is that $r_\rho$ still depends on the
unknown density. CTEM removes this unknown target algebraically and yields a
sample-based objective with a constant regression target.

\subsection{Constant-target energy matching}
\label{subsec:objective}

To specify which pairs of states are compared, let
$\omega:\mathcal{Z}\times\mathcal{Z}\to\R_{\ge0}$ be a non-negative comparison
weight. Throughout the main text, we assume that $\omega$ is symmetric,
$\omega(z,\tilde z)=\omega(\tilde z,z)$. The weight $\omega$ controls the
comparison geometry, such as local perturbations in continuous spaces or
neighborhood edges in discrete spaces.

We begin with the density-weighted matching objective
\begin{equation}
\mathcal{J}_\omega(\theta)
:=
\iint
\left[
    \tanh\!\left(
        \frac{f_\theta(z)-f_\theta(\tilde z)}{2}
    \right)
    -
    \frac{\rho(z)-\rho(\tilde z)}
         {\rho(z)+\rho(\tilde z)}
\right]^2
\bigl(\rho(z)+\rho(\tilde z)\bigr)
\omega(z,\tilde z)\,\nu(dz)\nu(d\tilde z).
\label{eq:oracle}
\end{equation}
This objective is not directly computable from samples, since it contains the
unknown density $\rho$ both in the modified ratio and in the density-weighted
measure. Nevertheless, its $\theta$-dependent part admits a Monte Carlo
estimable representation. The key observation is that the model transform
\[
    T_\theta(z,\tilde z)
    :=
    \tanh\!\left(
        \frac{f_\theta(z)-f_\theta(\tilde z)}{2}
    \right)
\]
is antisymmetric in $(z,\tilde z)$, whereas
$\omega(z,\tilde z)\nu(dz)\nu(d\tilde z)$ is symmetric. After expanding
\eqref{eq:oracle}, the terms involving $\rho(\tilde z)$ can therefore be
exchanged with those involving $\rho(z)$, so that all $\theta$-dependent terms
depend on $\rho$ only through the data measure $P(dz)=\rho(z)\nu(dz)$. This
gives a sample-based objective whose regression target is the constant $1$.

\begin{theorem}[Computable constant-target energy matching objective]
\label{thm:unified}
Suppose $\rho>0$ holds $\nu$-almost everywhere and
$\omega(z,\tilde z)=\omega(\tilde z,z)$. Then the objective
\eqref{eq:oracle} is equivalent, up to additive and multiplicative constants
independent of $\theta$, to the sample-only objective
\begin{equation}
\mathcal{L}_\omega(\theta)
=
\iint
\left[
    \tanh\!\left(
        \frac{f_\theta(z)-f_\theta(\tilde z)}{2}
    \right)
    -1
\right]^2
\omega(z,\tilde z)\,\rho(z)\,\nu(dz)\nu(d\tilde z).
\label{eq:sample-only}
\end{equation}
\end{theorem}

The proof, deferred to Appendix~\ref{app:proofs}, follows by expanding
\eqref{eq:oracle} and using the symmetry of
$\omega(z,\tilde z)\nu(dz)\nu(d\tilde z)$ together with the antisymmetry of
$T_\theta(z,\tilde z)$. This converts the density-weighted matching criterion
into the constant-target objective \eqref{eq:sample-only}.

Theorem~\ref{thm:unified} is the central reduction of CTEM. Although
\eqref{eq:oracle} contains the unknown density $\rho$, optimizing it is
equivalent to minimizing \eqref{eq:sample-only}, which only requires data
samples $z\sim P$ and comparison samples $\tilde z$ drawn according to the
weight $\omega(z,\tilde z)$.Thus CTEM avoids explicit
estimation of unstable density ratios and replaces them with a sample-based
energy matching loss with the constant target $1$. 

Recovery of $\log\rho$ up to an additive constant depends on whether the chosen
comparisons sufficiently connect the state space. We make this condition
explicit after specifying the concrete continuous and discrete training losses.The general formulation immediately covers product spaces, including mixed
continuous--discrete variables, by choosing $\mathcal{Z}$ and $\nu$
accordingly. For later use, we record the two most common specializations.
\begin{corollary}[Continuous variables]
\label{cor:continuous}
If $\mathcal{Z}\subseteq\R^d$, $\nu$ is the Lebesgue measure, $\rho=dP/d\nu>0$,
and $\omega$ is symmetric and non-negative, then the CTEM objective becomes
\begin{equation}
\mathcal{L}_{\omega,\mathrm{cont}}(\theta)
=
\iint
\left[
    \tanh\!\left(
        \frac{f_\theta(z)-f_\theta(\tilde z)}{2}
    \right)
    -1
\right]^2
\omega(z,\tilde z)\,\rho(z)\,dz\,d\tilde z.
\label{eq:cont-loss}
\end{equation}
\end{corollary}

\begin{corollary}[Discrete variables]
\label{cor:discrete}
If $\mathcal{Z}=\{z_1,\ldots,z_K\}$, $P(z_i)=p_i$, $\nu$ is the counting
measure, and $\omega_{ij}=\omega_{ji}\ge0$, then the CTEM objective becomes
\begin{equation}
\mathcal{L}_{\omega,\mathrm{disc}}(\theta)
=
\sum_{i=1}^K\sum_{j=1}^K
p_i\,\omega_{ij}
\left[
    \tanh\!\left(
        \frac{f_\theta(z_i)-f_\theta(z_j)}{2}
    \right)
    -1
\right]^2.
\label{eq:disc-loss}
\end{equation}
\end{corollary}

Both losses have the same sample-only structure and differ only in the
underlying reference measure and comparison weight. The continuous case will be
instantiated with local directional or Gaussian comparisons, while the discrete
case will be instantiated with graph-based neighborhood comparisons.
\subsection{Training loss for continuous variables}
\label{subsec:training-loss}

We now instantiate Corollary~\ref{cor:continuous} on
$\mathcal{Z}\subseteq\R^d$. Directly evaluating
\eqref{eq:cont-loss} requires integrating over all comparison points
$\tilde z$ for each data sample $z$, which is computationally prohibitive in
continuous spaces. The role of the comparison weight $\omega$ is therefore to
select tractable comparisons around each data point while preserving the same
energy-matching target. This induces a practical trade-off between
computational efficiency and the amount of comparison information used in each
update.

We consider two simple comparison weights controlled by a single scale
parameter $\varepsilon>0$. Both define, for each anchor $z$, a normalized
conditional distribution over comparison points $\tilde z$. The first compares
$z$ with points on a sphere of radius $2\varepsilon$, while the second uses an
isotropic Gaussian comparison kernel:
\begin{equation}
\omega_\varepsilon(z,\tilde z)
=
\begin{cases}
\displaystyle
\dfrac{1}{|\mathbb{S}^{d-1}|(2\varepsilon)^{d-1}}\,
\delta\!\left(\|z-\tilde z\|-2\varepsilon\right),
& \text{Spherical}, \\[2.0ex]
\displaystyle
\dfrac{1}{(8\pi\varepsilon^2)^{d/2}}\,
\exp\!\left(-\dfrac{\|z-\tilde z\|^2}{8\varepsilon^2}\right),
& \text{Gaussian}.
\end{cases}
\label{eq:omega-continuous-choices}
\end{equation}
Both choices are symmetric in $(z,\tilde z)$ and have $z$-independent
normalization. Hence they satisfy the symmetry requirement of
Theorem~\ref{thm:unified} and can be sampled directly without a Metropolis
correction.

Substituting \eqref{eq:omega-continuous-choices} into
Corollary~\ref{cor:continuous} yields two sampleable training objectives. Given
a mini-batch $\{z_b\}_{b=1}^B$ with $z_b\sim P$ and $M$ independent comparison
samples per anchor, we use
\begin{equation}
\widehat{\mathcal{L}}_{\varepsilon}(\theta)
=
\begin{cases}
\displaystyle
\frac{1}{BM}
\sum_{b=1}^B\sum_{m=1}^M
\left[
    \tanh\!\left(
        \frac{
        f_\theta(z_b)-f_\theta(z_b-2\varepsilon u_{b,m})
        }{2}
    \right)
    -1
\right]^2,
& \text{Spherical}, \\[4.0ex]
\displaystyle
\frac{1}{BM}
\sum_{b=1}^B\sum_{m=1}^M
\left[
    \tanh\!\left(
        \frac{
        f_\theta(z_b)-f_\theta(z_b-2\varepsilon \xi_{b,m})
        }{2}
    \right)
    -1
\right]^2,
& \text{Gaussian}.
\end{cases}
\label{eq:continuous-training-losses}
\end{equation}
Here $u_{b,m}\sim\Unif(\mathbb{S}^{d-1})$ for the spherical loss and
$\xi_{b,m}\sim\mathcal{N}(0,I_d)$ for the Gaussian loss.
We derive both forms in Appendix~\ref{app:continuous-loss}. In practice,
training only requires data samples $z_b\sim P$ and independent comparison
variables, given by either spherical directions $u_{b,m}$ or Gaussian
increments $\xi_{b,m}$. The parameter $\varepsilon$ controls the comparison scale. Small perturbations
provide local energy contrasts, whereas larger perturbations explore a wider
region of the data space and impose longer-range constraints. Importantly, in
contrast to denoising score matching, where the perturbation level directly
controls the bias of the learned score, CTEM identifies the same limiting
energy for any fixed $\varepsilon>0$ on each connected component induced by $\omega_\varepsilon$, provided the model
class is sufficiently expressive. Thus $\varepsilon$ affects the stability and
efficiency of training, but it does not change the limiting energy-matching
target.
\subsection{Training loss for discrete variables}
\label{subsec:training-loss-disc}

We now instantiate the discrete objective in \eqref{eq:disc-loss}. The
comparison weight $\omega_{ij}$ specifies which pairs of states are used to
estimate energy differences. For a product space $\mathcal{Z}=V^L$, two common comparison geometries are
local Hamming-one changes and global uniform corruptions:
\begin{equation}
    \omega_{\mathrm{loc}}(z,\tilde z)
    =
    \mathbf{1}\{d_H(z,\tilde z)=1\},
    \qquad
    \omega_{\mathrm{unif}}(z,\tilde z)
    =
    (1-\alpha)\mathbf{1}\{\tilde z=z\}
    +
    \alpha |V|^{-L},
    \quad \alpha\in[0,1].
    \label{eq:omega-discrete-examples}
\end{equation}
Here $d_H$ denotes Hamming distance. The first choice allows only
single-coordinate substitutions, whereas the second is the standard uniform
corruption kernel in discrete diffusion: it keeps the current state with
probability $1-\alpha$ and otherwise redraws a state uniformly from $V^L$.

Given a mini-batch $\{z_b\}_{b=1}^B$ with $z_b\sim P$, the discrete CTEM
training loss is
\begin{equation}
\widehat{\mathcal{L}}_{\mathrm{disc}}(\theta)
=
\frac{1}{BM}
\sum_{b=1}^B\sum_{m=1}^M
\left[
    \tanh\!\left(
        \frac{
        f_\theta(z_b)-f_\theta(\tilde z_{b,m})
        }{2}
    \right)
    -1
\right]^2,
\qquad
\tilde z_{b,m}\sim\omega(\cdot\mid z_b).
\label{eq:disc-training-loss}
\end{equation}
Here $\omega(\cdot\mid z)$ denotes the normalized comparison rule associated
with the pair weights $\omega_{ij}$. On each connected component where the
comparison graph is connected, the learned energy recovers $\log p_i$ up to an
additive constant.

For high-dimensional discrete data, we further apply CTEM to noisy states, as
in discrete diffusion models. Let
$q_{\sigma\mid 0}(z_\sigma\mid z_0)$ be a forward corruption kernel and let
$f_\theta(\cdot;\sigma)$ be a noise-conditional energy. Replacing clean data
states in \eqref{eq:disc-training-loss} with noisy states
$z_{\sigma,b}\sim q_{\sigma\mid 0}(\cdot\mid z_{0,b})$ gives a denoised CTEM
loss whose limiting target is
$f_\theta(\cdot;\sigma)=\log\rho_\sigma+C_\sigma$, where
$\rho_\sigma=dP_\sigma/d\nu$ is the density of the corrupted marginal. The
full expression and derivation are given in Appendix~\ref{app:denoised}.
\section{Experiments}
\label{sec:experiments}

\subsection{Continuous variables}
\label{subsec:exp-continuous}

\paragraph{Setup.}
We evaluate continuous CTEM on six density-estimation benchmarks: four
two-dimensional distributions, \emph{Spiral}, \emph{2-Gaussian}, \emph{Banana},
and \emph{Two Rings}, and two isotropic four-mode Gaussian mixtures in
$\R^{10}$ and $\R^{30}$. We compare with classical KDE baselines, including
Silverman KDE~\citep{silverman1986density} and cross-validated KDE
(CV-KDE)~\citep{bowman1984}, as well as learned density estimators, including
score-debiased KDE~\citep{epstein2025score} (SD-KDE), RealNVP
~\citep{dinh2017density}, and Neural Spline Flow (NSF)
~\citep{durkan2019neural}. We report density MSE for density recovery and
Fisher divergence for score recovery when the exact score is available.
Implementation details and metric definitions are given in
Appendix~\ref{app:exp-cont-setup}.

\begin{table}[htbp]
\centering
\caption{Continuous density estimation. Top block: 2-D density MSE
($\downarrow$, $\times 10^{-2}$, mean $\pm$ std over 5 seeds). Bottom block:
high-dimensional GMM Fisher divergence and density MSE. CTEM-S uses the
fixed-radius kernel and CTEM-G the Gaussian kernel. Best per row in
\textbf{bold}, second-best \underline{underlined}.}
\label{tab:continuous}
\small
\setlength{\tabcolsep}{3pt}
\renewcommand{\arraystretch}{1.05}
\resizebox{\linewidth}{!}{%
\begin{tabular}{l c ccccccc}
\toprule
Dataset & Metric & Silverman & CV-KDE & SD-KDE & RealNVP & NSF & \textbf{CTEM-S} & \textbf{CTEM-G} \\
\midrule
Spiral       & MSE & 0.883\,\tiny$\pm$0.03 & \underline{0.279}\,\tiny$\pm$0.03 & 0.429\,\tiny$\pm$0.03 & 0.952\,\tiny$\pm$0.23 & 0.694\,\tiny$\pm$0.07 & \textbf{0.256}\,\tiny$\pm$0.05 & 0.293\,\tiny$\pm$0.04 \\
2-Gaussian   & MSE & 0.880\,\tiny$\pm$0.08 & 0.381\,\tiny$\pm$0.08             & 0.251\,\tiny$\pm$0.07 & 4.530\,\tiny$\pm$0.29 & 2.065\,\tiny$\pm$0.40 & \underline{0.231}\,\tiny$\pm$0.13 & \textbf{0.223}\,\tiny$\pm$0.10 \\
Banana       & MSE & 0.476\,\tiny$\pm$0.01 & 0.222\,\tiny$\pm$0.10             & 0.154\,\tiny$\pm$0.01 & 0.984\,\tiny$\pm$0.22 & 0.468\,\tiny$\pm$0.08 & \underline{0.097}\,\tiny$\pm$0.02 & \textbf{0.081}\,\tiny$\pm$0.02 \\
Two Rings    & MSE & 8.633\,\tiny$\pm$0.10 & 2.002\,\tiny$\pm$0.10             & 5.596\,\tiny$\pm$0.10 & 7.281\,\tiny$\pm$2.03 & 2.660\,\tiny$\pm$0.43 & \underline{0.774}\,\tiny$\pm$0.39 & \textbf{0.773}\,\tiny$\pm$0.32 \\
\midrule
GMM (10-D)   & FD              & 4.39\,\tiny$\pm$0.02 & 1.85\,\tiny$\pm$0.04 & 4.35\,\tiny$\pm$0.02 & $(1.9\!\pm\!0.2){\times}10^{3}$ & $(2.2\!\pm\!0.5){\times}10^{4}$ & \underline{0.43}\,\tiny$\pm$0.08 & \textbf{0.39}\,\tiny$\pm$0.03 \\
             & MSE/$10^{-12}$  & 1.37\,\tiny$\pm$0.06 & 1.90\,\tiny$\pm$0.03 & 1.37\,\tiny$\pm$0.06 & 2.20\,\tiny$\pm$0.07            & $210$\,\tiny$\pm$\,66            & \underline{0.96}\,\tiny$\pm$0.21 & \textbf{0.29}\,\tiny$\pm$0.02 \\
\midrule
GMM (30-D)   & FD              & 25.02\,\tiny$\pm$0.15 & 10.27\,\tiny$\pm$0.04 & 25.02\,\tiny$\pm$0.15 & $(1.05\!\pm\!0.05){\times}10^{4}$ & $(1.2\!\pm\!0.2){\times}10^{5}$ & \underline{1.71}\,\tiny$\pm$0.05 & \textbf{1.50}\,\tiny$\pm$0.08 \\
             & MSE/$10^{-33}$  & 3.54\,\tiny$\pm$0.01  & 3.66\,\tiny$\pm$0.01  & 3.54\,\tiny$\pm$0.01  & 3.65\,\tiny$\pm$0.01              & 3.67\,\tiny$\pm$0.01            & \underline{0.99}\,\tiny$\pm$0.28 & \textbf{0.61}\,\tiny$\pm$0.05 \\
\bottomrule
\end{tabular}%
}
\end{table}

\begin{figure}[t]
\centering
\includegraphics[width=\linewidth]{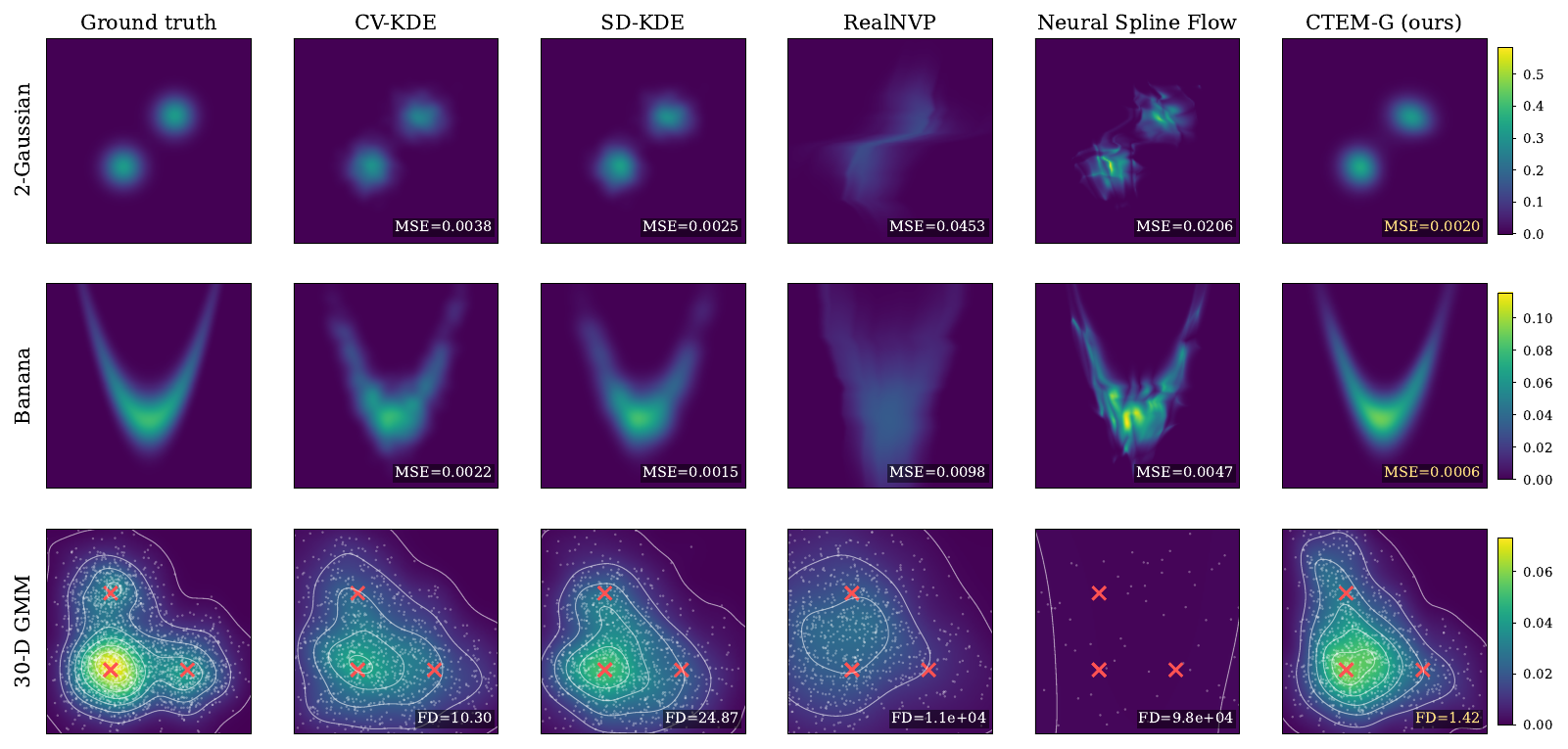}
\caption{Continuous density estimation. Top rows: learned densities on
2-Gaussian and Banana. Bottom row: 30-D GMM samples obtained by Langevin
dynamics using the learned score, projected onto $(x_1,x_2)$. Complementary
results are shown in Figure~\ref{fig:continuous-appendix}.}
\label{fig:continuous}
\end{figure}

\paragraph{Results.}
CTEM achieves the best density estimates across both low- and high-dimensional
settings (Table~\ref{tab:continuous}). On the 2-D benchmarks, CTEM attains the
lowest MSE on all four distributions. The gains are most pronounced on
geometrically structured or disconnected supports: CTEM-G improves over the
strongest baseline by $1.9{\times}$ on \emph{Banana} and $2.6{\times}$ on
\emph{Two Rings}. In high dimensions, CTEM also gives the lowest density MSE
and Fisher divergence on both GMMs; on the 30-D mixture, CTEM-G reduces Fisher
divergence by $7.3{\times}$ relative to the strongest baseline. The scale analysis in Appendix~\ref{app:gauss-vs-sphere} further shows that
CTEM does not require the perturbation scale $\varepsilon$ to vanish. Larger
comparison radii can cover a broader region of the data space and often yield
better approximation, consistent with the fact that the CTEM objective
identifies the same limiting energy for any fixed symmetric comparison kernel.

The learned energy also induces higher-quality score fields. As shown in
Appendix~\ref{app:fig-score}, CTEM-G closely matches the ground-truth score
magnitude and direction across the visualized benchmarks, while KDE methods
tend to oversmooth, RealNVP underestimates score magnitude, and NSF introduces
sharp spline-induced artifacts. The main discrepancy for CTEM occurs near
low-density boundaries, such as the outer region of \emph{Banana}, where the
training data provide limited information; this behavior is expected for
sample-based density estimation. Score-driven Langevin samples further confirm the quality of the
learned densities. Figure~\ref{fig:continuous} and
Appendix~\ref{app:fig-samples} show that CTEM samples concentrate on the
correct modes in both 2-D and high-dimensional mixtures, whereas baselines
either blur disconnected supports, miss modes, or produce diffuse samples.

\subsection{Discrete variables}
\label{subsec:exp-discrete}

\paragraph{Setup.}
We evaluate discrete CTEM on two tasks. First, we use quantized 2-D toy
distributions to test density recovery on sparse discrete supports. We compare
CTEM with Concrete Score Matching (CSM)~\citep{meng2022concrete} and
SEDD~\citep{lou2024discrete} under the same free per-state energy
parameterization, optimizer, comparison graph, and training budget, so that the
comparison isolates the effect of the training objective. Second, we evaluate
CTEM on binary MNIST to test whether the same constant-target objective scales
to high-dimensional discrete image generation.We report total variation (TV) distance and KL divergence for toy density
estimation, and classifier-based sample recognizability for MNIST. Experimental details are given in Appendix~\ref{app:exp-disc-setup}.

\begin{figure}[htbp]
\centering
\includegraphics[width=0.65\linewidth]{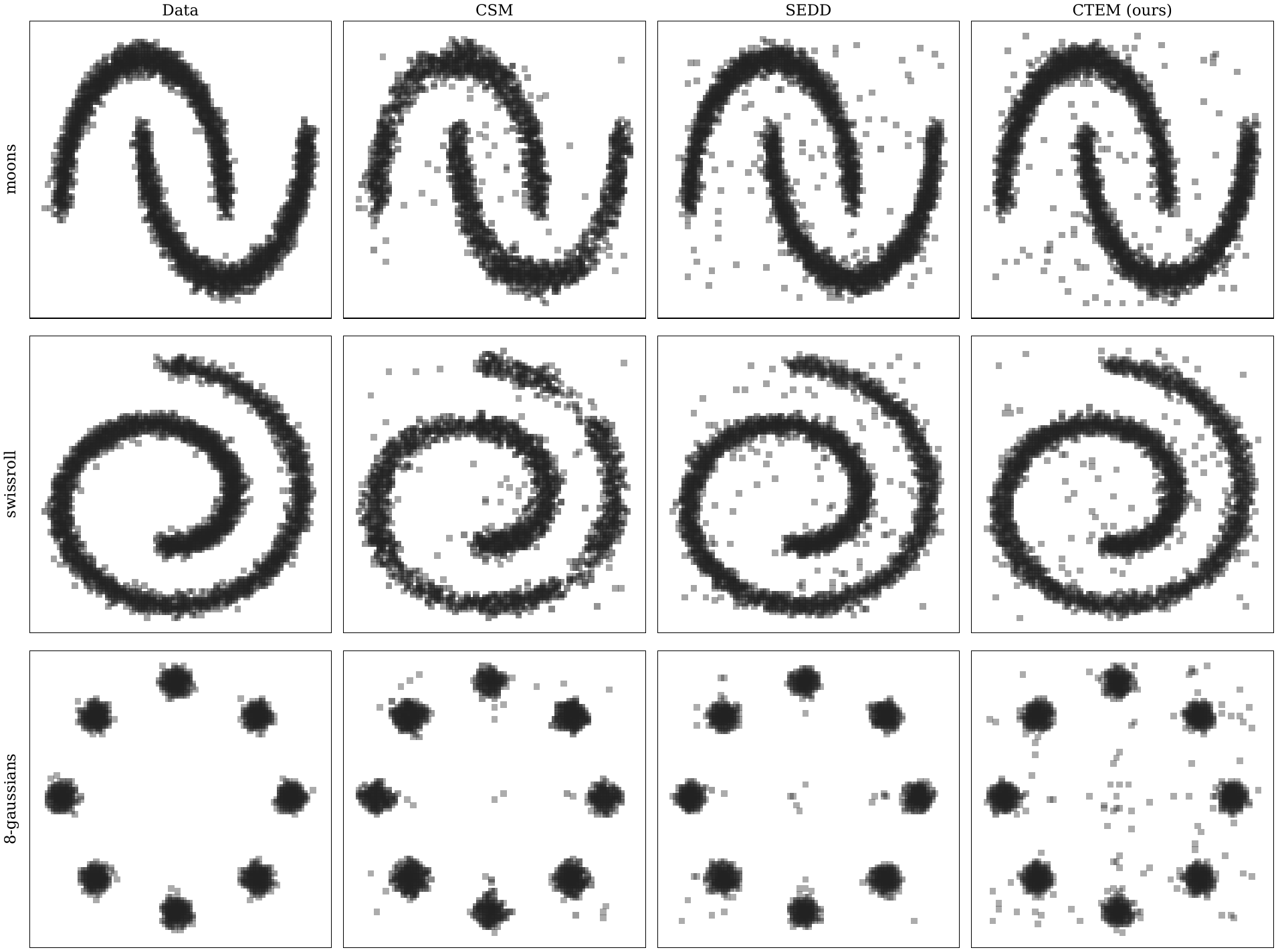}
\caption{Generated samples on quantized 2-D toy distributions. Rows
show moons, swissroll, and 8-Gaussians; columns show data, CSM, SEDD, and CTEM.}
\label{fig:disc-toy}
\end{figure}

\begin{table}[htbp]
\centering
\caption{Discrete density recovery on $91{\times}91$ quantized 2-D toy
distributions. TV distance and KL divergence are computed against the
ground-truth density; lower is better. CSM, SEDD, and CTEM (Sample) are
evaluated from sample-based density estimates, while CTEM ($p$) directly
evaluates the learned probability. Best values are in \textbf{bold} and
second-best values are \underline{underlined}.}
\label{tab:disc-toy}
\footnotesize
\setlength{\tabcolsep}{4pt}
\renewcommand{\arraystretch}{1.05}
\resizebox{\linewidth}{!}{%
\begin{tabular}{l cccc cccc}
\toprule
& \multicolumn{4}{c}{TV $\downarrow$}
& \multicolumn{4}{c}{KL $\downarrow$} \\
\cmidrule(lr){2-5}\cmidrule(lr){6-9}
Dataset
& CSM & SEDD & CTEM (Sample) & CTEM ($p$)
& CSM & SEDD & CTEM (Sample) & CTEM ($p$) \\
\midrule
2-D moons
& $0.387{\pm}0.021$
& $0.221{\pm}0.062$
& $\underline{0.218{\pm}0.061}$
& $\mathbf{0.216{\pm}0.062}$
& $3.466{\pm}0.426$
& $\mathbf{0.247{\pm}0.113}$
& $0.276{\pm}0.119$
& $\underline{0.251{\pm}0.115}$ \\
2-D swissroll
& $0.370{\pm}0.009$
& $0.172{\pm}0.018$
& $\underline{0.164{\pm}0.015}$
& $\mathbf{0.160{\pm}0.016}$
& $3.723{\pm}0.124$
& $\underline{0.139{\pm}0.023}$
& $0.157{\pm}0.022$
& $\mathbf{0.134{\pm}0.023}$ \\
2-D 8-Gaussians
& $0.328{\pm}0.073$
& $\mathbf{0.228{\pm}0.137}$
& $0.233{\pm}0.131$
& $\underline{0.231{\pm}0.133}$
& $1.192{\pm}0.559$
& $\underline{0.347{\pm}0.326}$
& $0.348{\pm}0.293$
& $\mathbf{0.306{\pm}0.269}$ \\
\bottomrule
\end{tabular}%
}
\end{table}

\begin{figure}[htbp]
\centering
\begin{minipage}{0.32\linewidth}
\centering
\includegraphics[width=\linewidth]{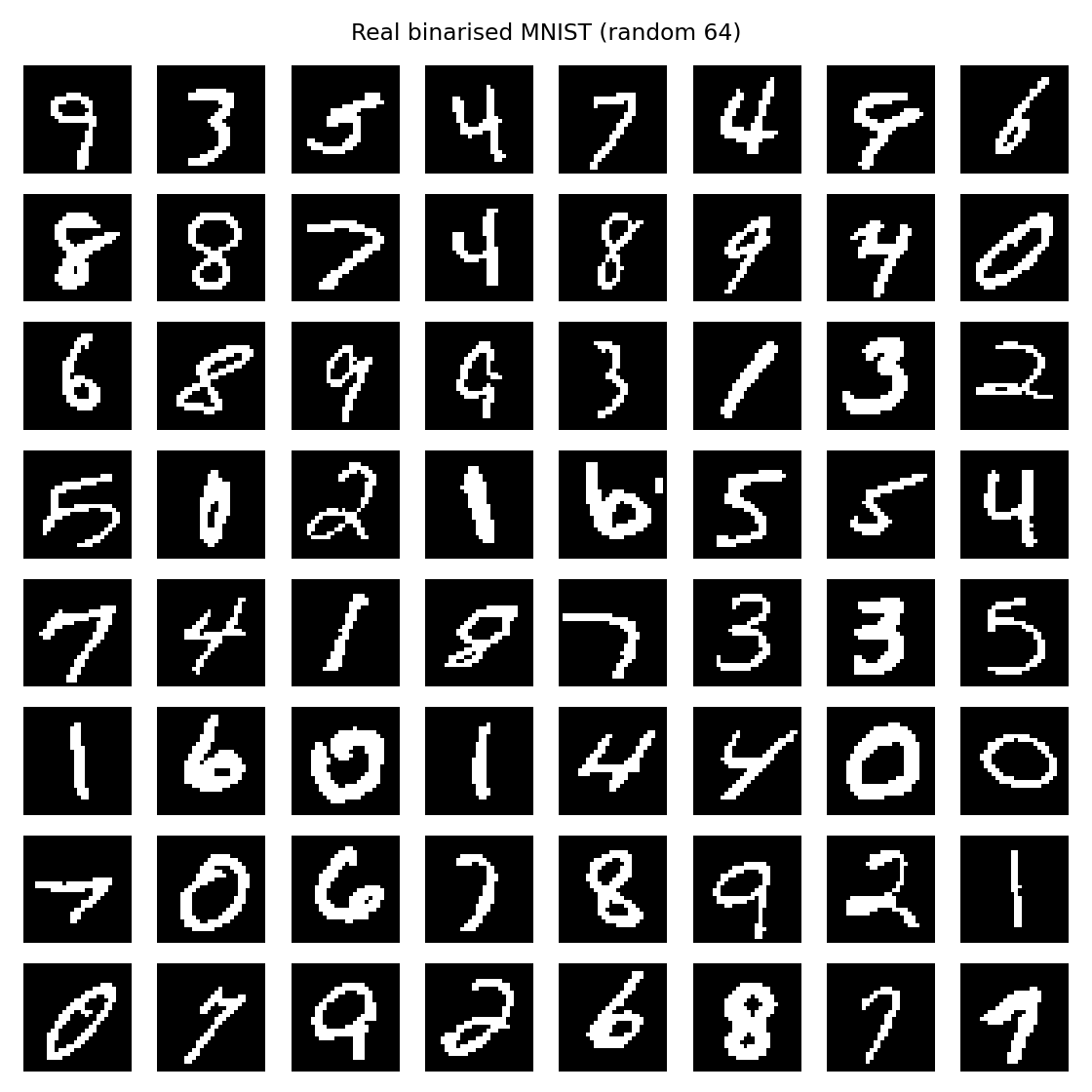}
\centerline{\small Real MNIST}
\end{minipage}
\hspace{0.08\linewidth}
\begin{minipage}{0.32\linewidth}
\centering
\includegraphics[width=\linewidth]{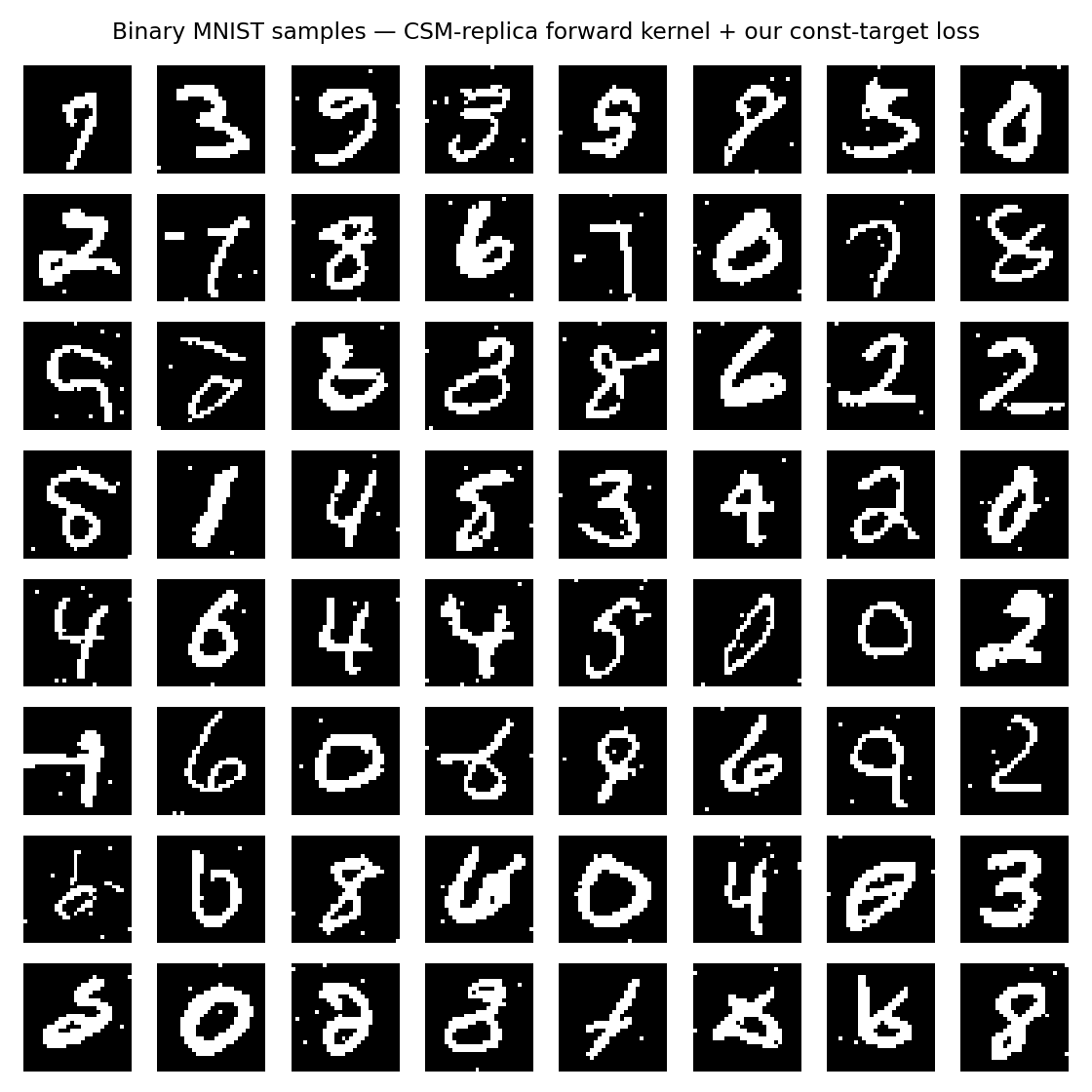}
\centerline{\small CTEM samples}
\end{minipage}
\caption{Binary MNIST samples. CTEM generates recognizable digit samples using
a noise-conditional energy and reverse-time Metropolis--Hastings sampling.}
\label{fig:mnist-samples}
\end{figure}

\paragraph{Results.}
Table~\ref{tab:disc-toy} and Figure~\ref{fig:disc-toy} show that CTEM recovers
the sparse discrete densities accurately across all three 2-D benchmarks. In
terms of TV distance, CTEM obtains the best result on moons and swissroll and
remains competitive with SEDD on 8-Gaussians. In terms of KL divergence, CTEM
achieves the best result on swissroll and 8-Gaussians, while SEDD is slightly
better on moons. A key distinction is that CSM and SEDD learn density
ratios, whereas CTEM learns a scalar energy whose normalized form directly
defines a distribution estimate $p$ on the state space. Overall,
CTEM ($p$) gives the best average performance across the three datasets, with
mean TV $0.202$ and mean KL $0.230$, compared with $0.207$ and $0.244$ for
SEDD.

On binary MNIST, CTEM produces visually recognizable samples
(Figure~\ref{fig:mnist-samples}). Using $1024$ generated samples, the
classifier top-1 recognizability score is $0.929$, close to real binary MNIST
($0.992$ under the same classifier). These results indicate that the
constant-target energy objective remains effective beyond small finite grids
and can be combined with standard discrete diffusion perturbations and
Metropolis--Hastings sampling for high-dimensional binary images.

\subsection{Mixed continuous--discrete product spaces}
\label{subsec:exp-mixed}

\paragraph{Setup.}
We finally evaluate CTEM on a mixed product space
$\mathcal{Z}=\mathbb{R}^2\times\{0,\ldots,K-1\}$ with $K=16$. The ground-truth
distribution is a labelled Gaussian mixture
\[
    p(x,y)=\pi_y\,p(x\mid y),
    \qquad
    p(x\mid y)=\mathcal{N}(x;\mu_y,I_2),
\]
where the component means $\{\mu_y\}_{y=0}^{K-1}$ lie on a ring and the label
prior $\pi_y$ is mildly non-uniform. We train an energy model
$f_\theta(x,y)$ with CTEM using a joint comparison kernel that perturbs the
continuous and discrete coordinates simultaneously. Additional implementation
details are given in Appendix~\ref{app:exp-mixed-setup}.

\begin{figure}[htbp]
\centering
\includegraphics[width=\linewidth]{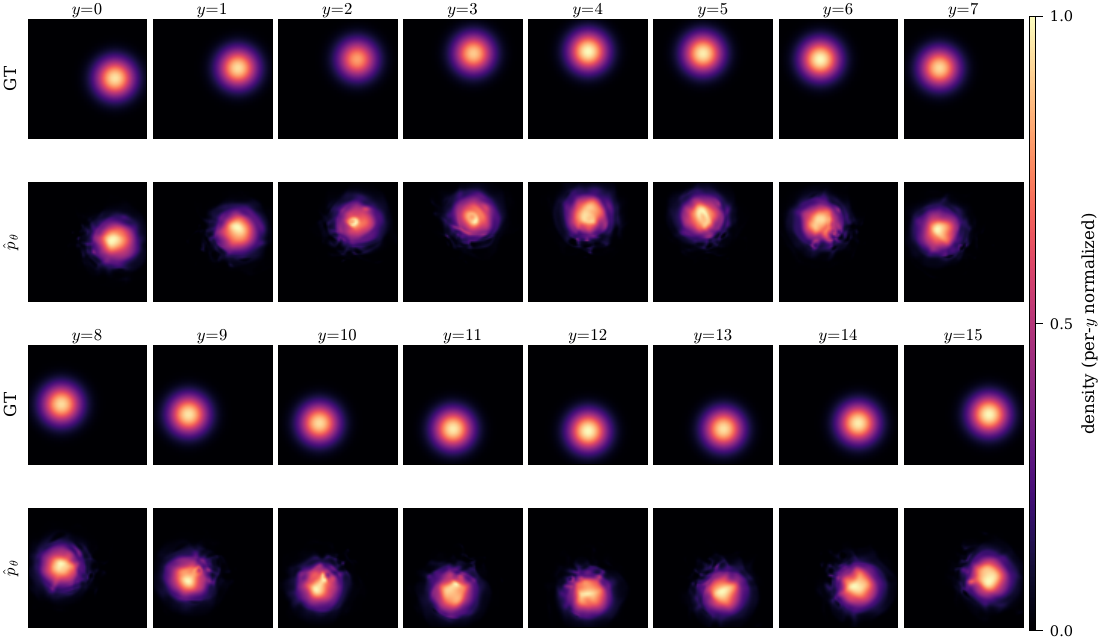}
\caption{Mixed continuous--discrete density estimation on
$\mathbb{R}^2\times\{0,\ldots,15\}$. Top row: ground-truth and learned
continuous marginals $p(x)$ and $\hat p_\theta(x)\propto\sum_y e^{f_\theta(x,y)}$.
Bottom panels: per-label joint components for each discrete state $y$. CTEM
recovers both the global mixed distribution and the conditional structure
associated with each discrete label.}
\label{fig:mixed-joint}
\end{figure}

\paragraph{Results.}
Figure~\ref{fig:mixed-joint} shows that CTEM accurately learns the mixed
distribution. In the top row, the learned marginal over the continuous
variable closely matches the ground-truth density and recovers the global
ring-shaped structure. In the bottom panels, the learned per-label components
also agree closely with the ground truth, showing that for each fixed discrete
state $y$, CTEM accurately captures the corresponding continuous conditional
distribution. These results indicate that a single scalar energy
$f_\theta(x,y)$ can successfully couple continuous and discrete coordinates and
recover a coherent mixed-state density.

\section{Related Work}
\label{sec:related}

\paragraph{Continuous density estimation.}
Classical density estimators include kernel methods
~\citep{rosenblatt1956remarks,silverman1986density,bowman1984},
energy-based models~\citep{lecun2006tutorial,du2019implicit}, and normalizing
flows~\citep{rezende2015variational,dinh2017density,papamakarios2017masked,
kingma2018glow,grathwohl2019ffjord,durkan2019neural}. Kernel methods are
sensitive to bandwidth and dimension, flows require invertible architectures,
and EBMs typically require partition-function or MCMC approximations. Score
matching avoids normalization by learning $\nabla\log\rho$
~\citep{hyvarinen2005estimation,vincent2011connection,song2019sliced}, forming
the basis of score-based diffusion models and related continuous generative
objectives~\citep{song2019generative,ho2020denoising,song2021score,
karras2022elucidating,lipman2023flow,liu2023rectified,song2023consistency}.
However, these methods rely on continuous differential structure, and denoising
objectives can be sensitive to the perturbation scale. CTEM instead estimates
the log-density through a bounded energy-difference transform. For any fixed
symmetric comparison rule that provides sufficient connectivity, the limiting
density-estimation target is unchanged; the comparison scale mainly affects
optimization stability and finite-sample efficiency. This contrasts with
denoising score objectives, where the perturbation level directly changes the
score-matching target.

\paragraph{Discrete diffusion models.}
Discrete score and diffusion methods replace continuous gradients with
finite-difference scores, probability ratios, or denoising transitions over
neighboring states~\citep{hyvarinen2007some,lyu2012interpretation,
austin2021structured,campbell2022continuous,meng2022concrete,lou2024discrete}.
Recent masked, absorbing-state, and flow-based variants further improve
scalability and language-modeling performance
~\citep{sahoo2024simple,shi2024simplified,ou2024absorbing,gat2024discrete,
sahoo2025diffusion}. These methods are effective, but their objectives are
usually tied to ordinary ratio targets, clean-data conditional distributions,
or a particular corruption/reverse-process parameterization. CTEM differs at
the objective level: it replaces unbounded ratio regression with a bounded
modified-ratio identity and yields a sample-only density-estimation objective
with the constant target $1$. The same constant-target energy-matching
principle applies to continuous, discrete, and mixed spaces, with the comparison
structure chosen according to the state space. This unified objective explains
the improved stability observed on sparse discrete supports and the accurate
density estimation in mixed continuous--discrete experiments.
\section{Conclusion}
\label{sec:conclusion}

We introduced \emph{Constant-Target Energy Matching} (CTEM), a unified
energy-based framework for density estimation on continuous, discrete, and
mixed state spaces. CTEM replaces the ordinary density-ratio target with a
bounded modified-ratio target and yields a sample-based density-estimation
objective with the constant target $1$ across heterogeneous domains. This formulation turns density estimation into a stable energy-matching problem
with a single objective across data types, avoiding the brittle ratio labels and
space-specific losses that often separate continuous and discrete generative
modeling. Empirically, CTEM improves density recovery, score
estimation, and sampling quality across continuous discrete and mixed continuous--discrete benchmarks.

This work also leaves several directions open. First, CTEM has not yet been
tested at the scale of modern language generation or large multimodal
generation, where model size, sequence length, and sampling efficiency become
central challenges. Second, while CTEM learns an energy function that can be
combined with standard sampling procedures, further work is needed to develop
faster and higher-quality samplers tailored to the constant-target objective.
Beyond these limitations, we believe that the same principle may provide a
useful route toward more stable score-function learning, and ultimately toward
a common training target for multimodal generative modeling across heterogeneous
data types.

\section*{Broader impacts}

This work proposes CTEM, a unified density-estimation framework for continuous,
discrete, and mixed-variable data. The paper is primarily methodological and is
evaluated on standard benchmark tasks, so it has no immediate direct societal
impact. Potential positive impacts include more stable probabilistic modeling
for structured and heterogeneous data. However, if used as a component of more
powerful generative models, CTEM may inherit standard risks of generative
modeling, such as synthetic-content misuse, privacy leakage, or biased
generation. Such applications should therefore follow standard precautions for
dataset governance, privacy, bias evaluation, and deployment safeguards.

{
\small
\bibliographystyle{plainnat}
\bibliography{references}
}


\appendix

\section{Proofs}
\label{app:proofs}

\subsection{Proof of Lemma~\ref{lem:identity}}

Let
\[
    a=\log\rho(z)-\log\rho(\tilde z).
\]
Since $\rho(z),\rho(\tilde z)>0$, we have
$e^a=\rho(z)/\rho(\tilde z)$. Using
$\tanh(a/2)=(e^a-1)/(e^a+1)$ gives
\begin{equation*}
\tanh(a/2)
=
\frac{\rho(z)/\rho(\tilde z)-1}
     {\rho(z)/\rho(\tilde z)+1}
=
\frac{\rho(z)-\rho(\tilde z)}
     {\rho(z)+\rho(\tilde z)}.
\end{equation*}
This proves the identity.

\subsection{Proof of Theorem~\ref{thm:unified}}

For compactness, define
\begin{equation*}
T_\theta(z,\tilde z)
:=
\tanh\!\left(
    \frac{f_\theta(z)-f_\theta(\tilde z)}{2}
\right).
\end{equation*}
Then $T_\theta$ is antisymmetric,
$T_\theta(\tilde z,z)=-T_\theta(z,\tilde z)$, while
$T_\theta^2$ is symmetric. Since
$\omega(z,\tilde z)=\omega(\tilde z,z)$, the measure
\[
    \omega(z,\tilde z)\,\nu(dz)\nu(d\tilde z)
\]
is invariant under the exchange of $z$ and $\tilde z$.

Expanding the integrand of \eqref{eq:oracle}, we obtain
\begin{align}
&
\left[
    T_\theta(z,\tilde z)
    -
    \frac{\rho(z)-\rho(\tilde z)}
         {\rho(z)+\rho(\tilde z)}
\right]^2
\bigl(\rho(z)+\rho(\tilde z)\bigr)
\nonumber\\
&\quad =
T_\theta(z,\tilde z)^2
\bigl(\rho(z)+\rho(\tilde z)\bigr)
-
2T_\theta(z,\tilde z)
\bigl(\rho(z)-\rho(\tilde z)\bigr)
+
\frac{
    \bigl(\rho(z)-\rho(\tilde z)\bigr)^2
}{
    \rho(z)+\rho(\tilde z)
}.
\label{eq:appendix-expand}
\end{align}
The last term is independent of $\theta$. After integration, it contributes
only the constant
\begin{equation*}
C_\rho
:=
\iint
\frac{
    \bigl(\rho(z)-\rho(\tilde z)\bigr)^2
}{
    \rho(z)+\rho(\tilde z)
}
\omega(z,\tilde z)\,\nu(dz)\nu(d\tilde z),
\end{equation*}
whenever the integral is finite.

We first simplify the quadratic term. By exchange symmetry and the symmetry of
$T_\theta^2$,
\begin{align}
&
\iint
T_\theta(z,\tilde z)^2
\rho(\tilde z)
\omega(z,\tilde z)\,\nu(dz)\nu(d\tilde z)
\nonumber\\
&\quad =
\iint
T_\theta(\tilde z,z)^2
\rho(z)
\omega(\tilde z,z)\,\nu(dz)\nu(d\tilde z)
\nonumber\\
&\quad =
\iint
T_\theta(z,\tilde z)^2
\rho(z)
\omega(z,\tilde z)\,\nu(dz)\nu(d\tilde z).
\label{eq:appendix-quad-swap}
\end{align}
Therefore,
\begin{align}
&
\iint
T_\theta(z,\tilde z)^2
\bigl(\rho(z)+\rho(\tilde z)\bigr)
\omega(z,\tilde z)\,\nu(dz)\nu(d\tilde z)
\nonumber\\
&\quad =
2
\iint
T_\theta(z,\tilde z)^2
\rho(z)
\omega(z,\tilde z)\,\nu(dz)\nu(d\tilde z).
\label{eq:appendix-quad}
\end{align}

We next simplify the linear term. By exchange symmetry and the antisymmetry of
$T_\theta$,
\begin{align}
&
\iint
T_\theta(z,\tilde z)
\rho(\tilde z)
\omega(z,\tilde z)\,\nu(dz)\nu(d\tilde z)
\nonumber\\
&\quad =
\iint
T_\theta(\tilde z,z)
\rho(z)
\omega(\tilde z,z)\,\nu(dz)\nu(d\tilde z)
\nonumber\\
&\quad =
-
\iint
T_\theta(z,\tilde z)
\rho(z)
\omega(z,\tilde z)\,\nu(dz)\nu(d\tilde z).
\label{eq:appendix-linear-swap}
\end{align}
Hence
\begin{align}
&
-2
\iint
T_\theta(z,\tilde z)
\bigl(\rho(z)-\rho(\tilde z)\bigr)
\omega(z,\tilde z)\,\nu(dz)\nu(d\tilde z)
\nonumber\\
&\quad =
-4
\iint
T_\theta(z,\tilde z)
\rho(z)
\omega(z,\tilde z)\,\nu(dz)\nu(d\tilde z).
\label{eq:appendix-linear}
\end{align}

Combining \eqref{eq:appendix-expand}, \eqref{eq:appendix-quad}, and
\eqref{eq:appendix-linear}, we obtain
\begin{align}
\mathcal{J}_\omega(\theta)
&=
2
\iint
\rho(z)
\left[
    T_\theta(z,\tilde z)^2
    -
    2T_\theta(z,\tilde z)
\right]
\omega(z,\tilde z)\,\nu(dz)\nu(d\tilde z)
+
C_\rho
\nonumber\\
&=
2
\iint
\rho(z)
\left[
    T_\theta(z,\tilde z)-1
\right]^2
\omega(z,\tilde z)\,\nu(dz)\nu(d\tilde z)
+
C_\rho',
\label{eq:appendix-J-to-L}
\end{align}
where
\begin{equation*}
C_\rho'
=
C_\rho
-
2
\iint
\rho(z)
\omega(z,\tilde z)\,\nu(dz)\nu(d\tilde z)
\end{equation*}
is independent of $\theta$. Since $P(dz)=\rho(z)\nu(dz)$, the
$\theta$-dependent term in \eqref{eq:appendix-J-to-L} is
\[
\iint
\left[
    T_\theta(z,\tilde z)-1
\right]^2
\omega(z,\tilde z)\,P(dz)\nu(d\tilde z),
\]
which is \eqref{eq:sample-only}. Thus
\[
    \mathcal{J}_\omega(\theta)
    =
    2\mathcal{L}_\omega(\theta)
    +
    C_\rho',
\]
where the multiplicative constant $2$ and additive constant $C_\rho'$ are
independent of $\theta$. Therefore minimizing
$\mathcal{J}_\omega$ is equivalent to minimizing
$\mathcal{L}_\omega$, proving the theorem.

\paragraph{Identifiability.}
Theorem~\ref{thm:unified} shows that the oracle density-matching objective and
the sample-based constant-target objective have the same minimizers. By Lemma~\ref{lem:identity} and the injectivity of
$\tanh$, 
\begin{equation}
    f_\theta(z)-f_\theta(\tilde z)
    =
    \log\rho(z)-\log\rho(\tilde z).
\label{eq:appendix-identifiability-difference}
\end{equation}
Equivalently,
\begin{equation}
    f_\theta(z)-\log\rho(z)
    =
    f_\theta(\tilde z)-\log\rho(\tilde z).
\label{eq:appendix-identifiability-constant}
\end{equation}
Thus both sides must equal the same constant along every admissible comparison,
so $f_\theta-\log\rho$ is constant on each connected component induced by
$\omega$. If the comparisons connect the relevant state space, then
$f_\theta(z)=\log\rho(z)+C$ on that component.

\subsection{Direct corollaries}
\label{app:direct-corollaries}

\paragraph{Continuous variables.}
Corollary~\ref{cor:continuous} follows by taking
$\mathcal{Z}\subseteq\R^d$ and choosing $\nu$ as the Lebesgue measure. Then
$P(dz)=\rho(z)dz$, and \eqref{eq:sample-only} becomes
\begin{equation*}
\mathcal{L}_{\omega,\mathrm{cont}}(\theta)
=
\iint
\left[
    \tanh\!\left(
        \frac{f_\theta(z)-f_\theta(\tilde z)}{2}
    \right)
    -1
\right]^2
\omega(z,\tilde z)\rho(z)\,dz\,d\tilde z,
\end{equation*}
which is exactly \eqref{eq:cont-loss}.

\paragraph{Discrete variables.}
Corollary~\ref{cor:discrete} follows by taking
$\mathcal{Z}=\{z_1,\ldots,z_K\}$ and choosing $\nu$ as the counting measure.
Then $\rho(z_i)=P(z_i)=p_i$. Substituting this into
\eqref{eq:sample-only} gives
\begin{align*}
\mathcal{L}_{\omega,\mathrm{disc}}(\theta)
&=
\sum_{i=1}^K\sum_{j=1}^K
p_i\,\omega_{ij}
\left[
    \tanh\!\left(
        \frac{f_\theta(z_i)-f_\theta(z_j)}{2}
    \right)
    -1
\right]^2,
\end{align*}
which is \eqref{eq:disc-loss}. The identifiability argument above further
implies that, on each connected component of the graph with edges
$\omega_{ij}>0$, the learned function satisfies
\[
    f_\theta(z_i)=\log p_i+C.
\]

\subsection{Derivation of the continuous training losses}
\label{app:continuous-loss}

We derive the two Monte Carlo objectives in
\eqref{eq:continuous-training-losses} from the continuous CTEM objective
\eqref{eq:cont-loss}. Throughout this section, define
\begin{equation}
g_\theta(z,\tilde z)
:=
\left[
    \tanh\!\left(
        \frac{f_\theta(z)-f_\theta(\tilde z)}{2}
    \right)
    -1
\right]^2 .
\label{eq:appendix-g-theta}
\end{equation}
Then
\[
    \mathcal{L}_{\omega,\mathrm{cont}}(\theta)
    =
    \int \rho(z)
    \left[
        \int
        g_\theta(z,\tilde z)\omega(z,\tilde z)\,d\tilde z
    \right] dz .
\]
Thus the practical objective is obtained by choosing a symmetric comparison
weight whose inner integral can be written as an expectation over a simple
random variable.

\paragraph{Fixed-radius comparisons.}
For the spherical comparison weight,
\begin{equation}
\omega_\varepsilon^{\mathrm{sphere}}(z,\tilde z)
=
\frac{1}{|\mathbb{S}^{d-1}|(2\varepsilon)^{d-1}}\,
\delta\!\left(\|z-\tilde z\|-2\varepsilon\right),
\label{eq:appendix-sphere-weight}
\end{equation}
the comparison point lies on the sphere of radius $2\varepsilon$ centered at
$z$. Using the polar change of variables
\[
    \tilde z=z-r u,
    \qquad
    r>0,\quad u\in\mathbb{S}^{d-1},
\]
we have $d\tilde z=r^{d-1}dr\,d\sigma(u)$, where $d\sigma$ is the surface
measure on $\mathbb{S}^{d-1}$. Therefore,
\begin{align}
&
\int
g_\theta(z,\tilde z)
\omega_\varepsilon^{\mathrm{sphere}}(z,\tilde z)
\,d\tilde z
\nonumber\\
&\quad =
\frac{1}{|\mathbb{S}^{d-1}|(2\varepsilon)^{d-1}}
\int_0^\infty
\int_{\mathbb{S}^{d-1}}
g_\theta(z,z-r u)
\delta(r-2\varepsilon)
r^{d-1}\,d\sigma(u)\,dr
\nonumber\\
&\quad =
\frac{1}{|\mathbb{S}^{d-1}|}
\int_{\mathbb{S}^{d-1}}
g_\theta(z,z-2\varepsilon u)\,d\sigma(u)
\nonumber\\
&\quad =
\E_{u\sim\Unif(\mathbb{S}^{d-1})}
\left[
    g_\theta(z,z-2\varepsilon u)
\right].
\label{eq:appendix-fixed-radius-reduction}
\end{align}
Substituting this identity into \eqref{eq:cont-loss} yields
\begin{equation}
\mathcal{L}_{\varepsilon}^{\mathrm{sphere}}(\theta)
=
\E_{z\sim P}
\E_{u\sim\Unif(\mathbb{S}^{d-1})}
\left[
    \tanh\!\left(
        \frac{
            f_\theta(z)-f_\theta(z-2\varepsilon u)
        }{2}
    \right)
    -1
\right]^2 .
\label{eq:appendix-sphere-loss}
\end{equation}
The empirical estimator with $B$ data samples and $M$ directions per sample is
the spherical case of \eqref{eq:continuous-training-losses}.

\paragraph{Gaussian comparisons.}
For the Gaussian comparison weight,
\begin{equation}
\omega_\varepsilon^{\mathrm{gauss}}(z,\tilde z)
=
\frac{1}{(8\pi\varepsilon^2)^{d/2}}
\exp\!\left(
    -\frac{\|z-\tilde z\|^2}{8\varepsilon^2}
\right),
\label{eq:appendix-gaussian-weight}
\end{equation}
the inner integral in \eqref{eq:cont-loss} is
\begin{equation}
\int
g_\theta(z,\tilde z)
\frac{1}{(8\pi\varepsilon^2)^{d/2}}
\exp\!\left(
    -\frac{\|z-\tilde z\|^2}{8\varepsilon^2}
\right)
d\tilde z .
\label{eq:appendix-gaussian-inner}
\end{equation}
Let
\[
    \tilde z=z-2\varepsilon\xi,
    \qquad
    \xi\sim\mathcal{N}(0,I_d).
\]
Then $d\tilde z=(2\varepsilon)^d d\xi$, and
\begin{equation}
\frac{1}{(8\pi\varepsilon^2)^{d/2}}
\exp\!\left(
    -\frac{\|z-\tilde z\|^2}{8\varepsilon^2}
\right)d\tilde z
=
\frac{1}{(2\pi)^{d/2}}
\exp\!\left(
    -\frac{\|\xi\|^2}{2}
\right)d\xi .
\label{eq:appendix-gaussian-change}
\end{equation}
Thus
\begin{align}
&
\int
g_\theta(z,\tilde z)
\omega_\varepsilon^{\mathrm{gauss}}(z,\tilde z)
\,d\tilde z
\nonumber\\
&\quad =
\E_{\xi\sim\mathcal{N}(0,I_d)}
\left[
    g_\theta(z,z-2\varepsilon\xi)
\right].
\label{eq:appendix-gaussian-reduction}
\end{align}
Substituting this identity into \eqref{eq:cont-loss} gives
\begin{equation}
\mathcal{L}_{\varepsilon}^{\mathrm{gauss}}(\theta)
=
\E_{z\sim P}
\E_{\xi\sim\mathcal{N}(0,I_d)}
\left[
    \tanh\!\left(
        \frac{
            f_\theta(z)-f_\theta(z-2\varepsilon \xi)
        }{2}
    \right)
    -1
\right]^2 .
\label{eq:appendix-gaussian-loss}
\end{equation}
The corresponding empirical estimator is the Gaussian case of
\eqref{eq:continuous-training-losses}.

\subsection{Noise-conditional discrete CTEM}
\label{app:denoised}

We derive the noise-conditional discrete CTEM objective by applying the clean
discrete objective to the corrupted marginal at each noise level. Let
$\mathcal{Z}=\{z_1,\ldots,z_K\}$ and let
$q_{\sigma\mid 0}(z_\sigma\mid z_0)$ be a forward corruption kernel on
$\mathcal{Z}$. The corrupted marginal is
\begin{equation}
    P_\sigma(z)
    =
    \sum_{z_0\in\mathcal{Z}}
    P(z_0)q_{\sigma\mid 0}(z\mid z_0).
    \label{eq:appendix-corrupted-marginal}
\end{equation}
Under the counting-measure convention used in the main text,
$\rho_\sigma(z_i)=P_\sigma(z_i)$.

We take the comparison rule $\omega(\cdot\mid z_i)$ to be a symmetric Markov
kernel,
\begin{equation}
    \omega_{ij}=\omega_{ji},
    \qquad
    \sum_{j=1}^K \omega_{ij}=1.
    \label{eq:appendix-symmetric-markov-kernel}
\end{equation}
This convention matches standard discrete transition kernels and is satisfied
by the normalized Hamming-one kernel on regular product spaces and by uniform
corruption kernels.

For a fixed noise level $\sigma$, applying
Corollary~\ref{cor:discrete} to the corrupted marginal $P_\sigma$ gives
\begin{equation}
\mathcal{L}_{\sigma}(\theta)
=
\sum_{i=1}^K\sum_{j=1}^K
P_\sigma(z_i)\,\omega_{ij}
\left[
    \tanh\!\left(
        \frac{
            f_\theta(z_i;\sigma)-f_\theta(z_j;\sigma)
        }{2}
    \right)
    -1
\right]^2 .
\label{eq:appendix-denoised-discrete-sum}
\end{equation}
Equivalently,
\begin{align}
\mathcal{L}_{\sigma}(\theta)
&=
\E_{z_\sigma\sim P_\sigma}
\E_{\tilde z\sim\omega(\cdot\mid z_\sigma)}
\left[
    \tanh\!\left(
        \frac{
            f_\theta(z_\sigma;\sigma)
            -
            f_\theta(\tilde z;\sigma)
        }{2}
    \right)
    -1
\right]^2 .
\label{eq:appendix-denoised-expectation}
\end{align}
A sample from $P_\sigma$ is obtained by drawing
$z_0\sim P$ and then
$z_\sigma\sim q_{\sigma\mid 0}(\cdot\mid z_0)$. Hence, given a mini-batch
$\{z_{0,b}\}_{b=1}^B$, an unbiased Monte Carlo estimator is
\begin{equation}
\widehat{\mathcal{L}}_{\sigma}(\theta)
=
\frac{1}{BM}
\sum_{b=1}^B\sum_{m=1}^M
\left[
    \tanh\!\left(
        \frac{
            f_\theta(z_{\sigma,b};\sigma)
            -
            f_\theta(\tilde z_{b,m};\sigma)
        }{2}
    \right)
    -1
\right]^2,
\label{eq:appendix-fixed-sigma-mc}
\end{equation}
where
\[
    z_{\sigma,b}\sim q_{\sigma\mid 0}(\cdot\mid z_{0,b}),
    \qquad
    \tilde z_{b,m}\sim\omega(\cdot\mid z_{\sigma,b}).
\]

Averaging over a noise schedule $\pi(\sigma)$ gives
\begin{align}
\mathcal{L}_{\mathrm{noise}}(\theta)
&=
\E_{\sigma\sim\pi}
\E_{z_0\sim P}
\E_{z_\sigma\sim q_{\sigma\mid 0}(\cdot\mid z_0)}
\E_{\tilde z\sim\omega(\cdot\mid z_\sigma)}
\nonumber\\
&\quad
\left[
    \tanh\!\left(
        \frac{
            f_\theta(z_\sigma;\sigma)
            -
            f_\theta(\tilde z;\sigma)
        }{2}
    \right)
    -1
\right]^2 .
\label{eq:appendix-denoised-loss-expectation}
\end{align}
For each fixed $\sigma$, the recovered log-density satisfies
\begin{equation}
    f_\theta(z_i;\sigma)
    =
    \log P_\sigma(z_i)+C_\sigma
    =
    \log\rho_\sigma(z_i)+C_\sigma
\end{equation}
on every connected component induced by the nonzero entries of $\omega$.
When $\sigma=0$ and the forward corruption kernel is the identity,
$P_\sigma=P$, and the objective reduces to the clean discrete CTEM objective.

\section{Compute resources}
\label{app:compute_re}
All experiments were run on a workstation with a single NVIDIA RTX 4090 GPU.
The 2-D continuous benchmark takes less than 10 minutes to run, and the
high-dimensional GMM experiments take less than 30 minutes. The 2-D discrete
density-estimation experiments take approximately 10 minutes, while the binary
MNIST experiment takes approximately 6 hours.

\section{Existing assets and licenses}
We use MNIST\cite{lecun1998mnist} for the binary image experiments and cite the original dataset.
For baseline comparisons, we use or adapt publicly available implementations
where applicable, including the open-source implementation of SEDD
~\citep{lou2024discrete}. We credit the original authors, follow the
corresponding licenses and terms of use, and report implementation details
needed to reproduce the comparisons.

\section{Continuous-variable experimental setup}
\label{app:exp-cont-setup}

This section provides the experimental details for
Section~\ref{subsec:exp-continuous}.

\paragraph{Datasets.}
We evaluate on four 2-D distributions and two high-dimensional Gaussian
mixtures. The 2-D datasets are: \emph{Spiral}, a noisy two-turn Archimedean
spiral; \emph{2-Gaussian}, an equal mixture of two isotropic Gaussians centered
at $(1,1)$ and $(-1,-1)$; \emph{Banana}, a Rosenbrock-style curved
distribution with $x_1\sim\mathcal{N}(0,4)$ and
$x_2=0.5(x_1^2-4)+\eta$, $\eta\sim\mathcal{N}(0,1)$; and \emph{Two Rings}, an
equal mixture of two annular distributions with radii $1$ and $2$. The
high-dimensional benchmarks are four-mode isotropic Gaussian mixtures in
$\R^{10}$ and $\R^{30}$, with unit covariance and component means
$3e_1,3e_2,3e_3,3e_4$. Closed-form densities and scores are available for all
benchmarks except the Spiral score.

For 2-D experiments, we use $n_\mathrm{train}=1000$ samples and evaluate density
MSE on a $200\times200$ grid. For high-dimensional experiments, we use
$n_\mathrm{train}=5000$ training samples and evaluate on $2000$ independent test
samples.

\paragraph{CTEM model.}
The energy network $f_\theta:\R^d\to\R$ is a three-hidden-layer MLP with SiLU
activations, width $128$ for 2-D experiments and width $256$ for the 10-D and
30-D GMMs. We train with Adam, batch size $256$, and $M=4$ antithetic
comparison directions per data point. The learning rate is $10^{-4}$ unless
otherwise selected in the Gaussian-kernel comparison. We train for $15$k steps
in 2-D and $30$k steps in high dimensions. The comparison scale is
$\varepsilon=c\bar\sigma$, where $\bar\sigma$ is the mean empirical
per-coordinate standard deviation. The fixed-radius kernel uses
$c=0.50$ for Spiral and Two Rings, $c=0.75$ for 2-Gaussian and Banana,
$c=0.08$ for the 10-D GMM, and $c=0.20$ for the 30-D GMM.

After training, scores are obtained by automatic differentiation of
$f_\theta$. For 2-D density evaluation, we normalize $\exp(f_\theta)$ on the
evaluation grid by trapezoidal integration. For high-dimensional density MSE,
the unknown normalizing constant of CTEM is estimated once by importance
sampling.

\paragraph{Baselines.}
Silverman KDE uses the standard rule-of-thumb bandwidth
$h=0.9\bar\sigma n^{-1/(d+4)}$~\citep{silverman1986density}. CV-KDE selects the
bandwidth by leave-one-out likelihood for $n\le500$ and 5-fold
cross-validation otherwise~\citep{bowman1984}. SD-KDE follows
\citet{epstein2025score}: it estimates a KDE score at the training samples,
shifts the samples by a score-debiasing correction, and reruns KDE on the
shifted set.

RealNVP~\citep{dinh2017density} and NSF~\citep{durkan2019neural} are
implemented with \texttt{nflows}. Both use standardized inputs, a standard
Gaussian base distribution, cosine learning-rate decay, small Gaussian
dequantization noise, gradient clipping, and weight decay. In 2-D, we use
8 coupling blocks and 10k optimization steps; in high dimensions, we use
10 coupling blocks and 40k steps. NSF uses rational-quadratic spline couplings
with linear tails.

\paragraph{Metrics.}
For 2-D distributions, density MSE is computed on a regular grid as a
trapezoidal approximation to $\int(\hat\rho(x)-\rho(x))^2dx$. For
high-dimensional GMMs, Fisher divergence is estimated as
\[
    \frac{1}{n_\mathrm{test}}
    \sum_{i=1}^{n_\mathrm{test}}
    \|\hat s(x_i)-\nabla\log\rho(x_i)\|^2,
    \qquad x_i\sim\rho.
\]
High-dimensional density MSE is evaluated on the same test samples as
$\frac{1}{n_\mathrm{test}}\sum_i(\hat\rho(x_i)-\rho(x_i))^2$. For 2-D results,
we report mean and standard deviation over five seeds.

\paragraph{Sampling visualization.}
For Langevin visualizations, we initialize $1000$ chains from
$\mathcal{N}(0,2^2I)$ and run unadjusted Langevin dynamics with step size
$5\times10^{-3}$. We use 2000 steps for the 2-D sample visualizations and
3000 steps for the high-dimensional projections in the main figure.

\begin{figure}[t]
\centering
\includegraphics[width=\linewidth]{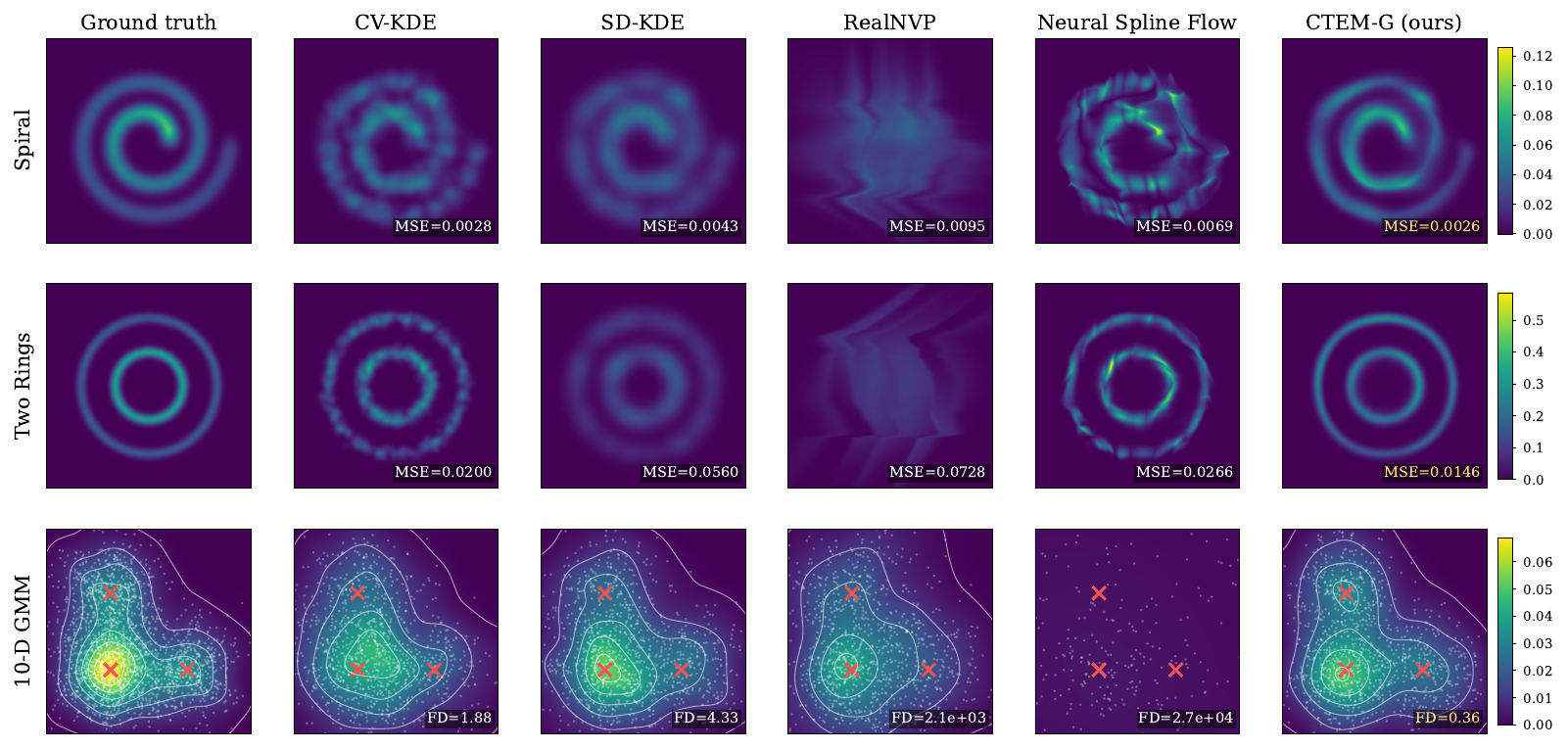}
\caption{Additional continuous-variable results. Top rows: learned densities
on Spiral and Two Rings. Bottom row: 10-D GMM samples obtained by Langevin
dynamics using the learned score, projected onto $(x_1,x_2)$.}
\label{fig:continuous-appendix}
\end{figure}

\section{Effect of comparison scale}
\label{app:gauss-vs-sphere}

The continuous CTEM loss in \eqref{eq:continuous-training-losses} supports both
fixed-radius comparisons, $z-2\varepsilon u$ with
$u\sim\Unif(\mathbb{S}^{d-1})$, and Gaussian comparisons,
$z-2\varepsilon\xi$ with $\xi\sim\mathcal{N}(0,I_d)$. Both kernels satisfy the
symmetry condition in Theorem~\ref{thm:unified} and identify the same limiting
energy. They differ mainly in their effective comparison radius: the Gaussian
increment has typical norm $2\varepsilon\sqrt d$, whereas the fixed-radius
increment has norm $2\varepsilon$. Thus, matching the two comparison scales in
high dimension requires the Gaussian $\varepsilon$ to be smaller by roughly a
factor of $\sqrt d$.

For the CTEM-G results in Table~\ref{tab:continuous}, we select
$\varepsilon=c\bar\sigma$ from a 7-point geometric grid and tune the learning
rate over three values, using the same architecture and optimization budget as
CTEM-S. The selected values of $c$ are
$\{0.30,1.00,1.50,0.30\}$ for Spiral, 2-Gaussian, Banana, and Two Rings, and
$\{0.02,0.04\}$ for the 10-D and 30-D GMMs. The smaller high-dimensional
Gaussian scales are consistent with the $1/\sqrt d$ effective-radius
correction.

\begin{figure}[h]
\centering
\includegraphics[width=\linewidth]{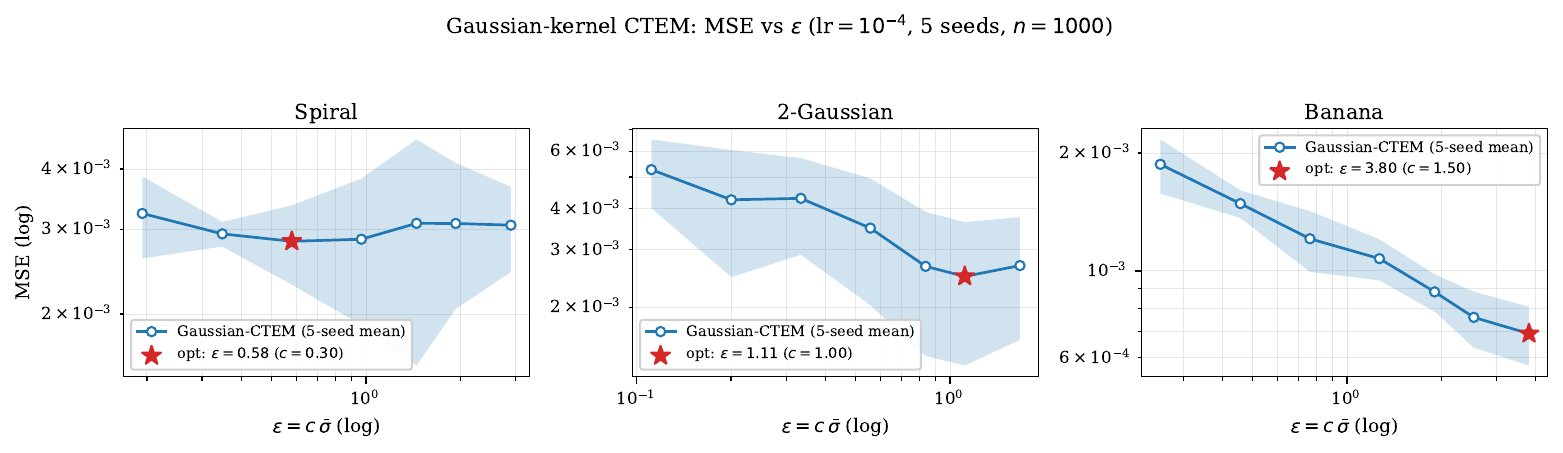}
\caption{Effect of the Gaussian comparison scale
$\varepsilon=c\bar\sigma$ on 2-D density recovery. CTEM remains stable over a
wide range of scales and often benefits from comparisons that cover a larger
region of the data space.}
\label{fig:eps-sweep}
\end{figure}

Figure~\ref{fig:eps-sweep} studies the effect of $\varepsilon$ for CTEM-G on
three 2-D benchmarks. The results show that CTEM does not require the
comparison scale to be infinitesimal. In fact, larger values of $\varepsilon$
often improve density recovery by comparing each data point with a broader
region of the state space. On Banana, the best value lies at the largest
tested scale, indicating that broad comparisons can be beneficial when the
support is curved and sparsely sampled. This behavior is consistent with the
constant-target objective: unlike denoising score matching, CTEM does not rely
on a small-noise asymptotic target.

\section{Score-field visualization}
\label{app:fig-score}

\begin{figure}[t]
\centering
\includegraphics[width=\linewidth]{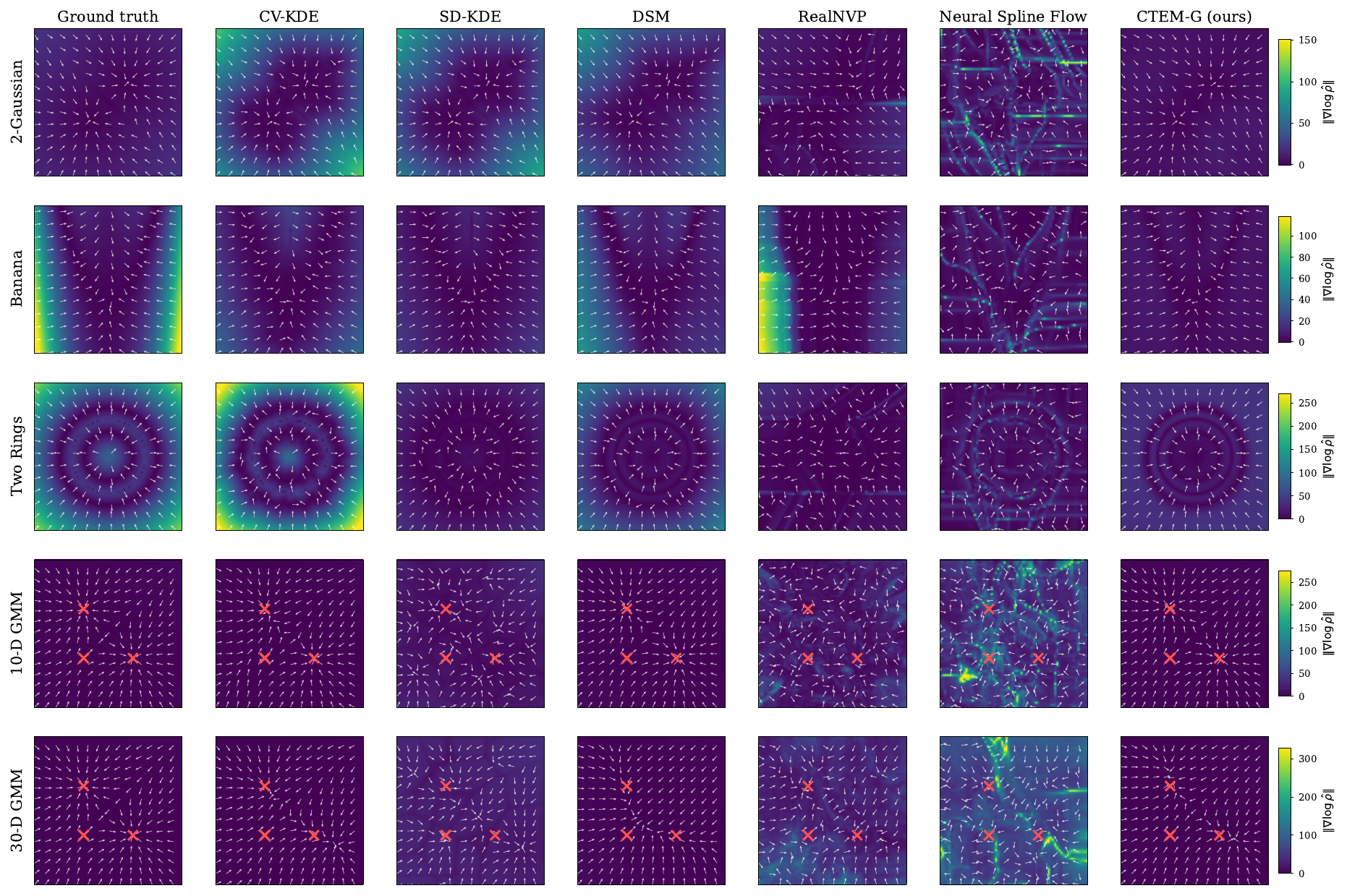}
\caption{Estimated score fields. Color shows
$\|\nabla\log\hat\rho(x)\|$ and arrows show normalized score directions. Top
rows: 2-D benchmarks. Bottom rows: slices of the 10-D and 30-D GMMs.}
\label{fig:continuous-score}
\end{figure}

Figure~\ref{fig:continuous-score} compares the score fields induced by each
density estimator. For 2-D benchmarks, we visualize
$\hat s(x)=\nabla\log\hat\rho(x)$ on the evaluation grid. For the 10-D and
30-D GMMs, we visualize the slice $x_{3:D}=0$, which contains two of the four
modes. We additionally include a denoising score matching (DSM) baseline
trained at noise level $\sigma=0.1$.

CTEM-G gives the closest match to the ground-truth score across both magnitude
and direction. KDE-based methods produce smooth but contracted scores, RealNVP
often underestimates the score magnitude, and NSF exhibits high-magnitude
artifacts near spline boundaries. CTEM is less accurate near low-density
boundaries, most visibly on Banana, where few training samples are available.
This is consistent with the sample-based nature of the estimator.

\section{Score-driven Langevin sampling}
\label{app:fig-samples}

We further evaluate the learned densities by using their scores in unadjusted
Langevin dynamics. For 2-D benchmarks, we initialize $1000$ chains from
$\mathcal{N}(0,2^2I)$ and run 2000 steps with step size $5\times10^{-3}$. For
the high-dimensional GMMs, we use the same initialization and visualize the
projection onto $(x_1,x_2)$.

\begin{figure}[t]
\centering
\includegraphics[width=\linewidth]{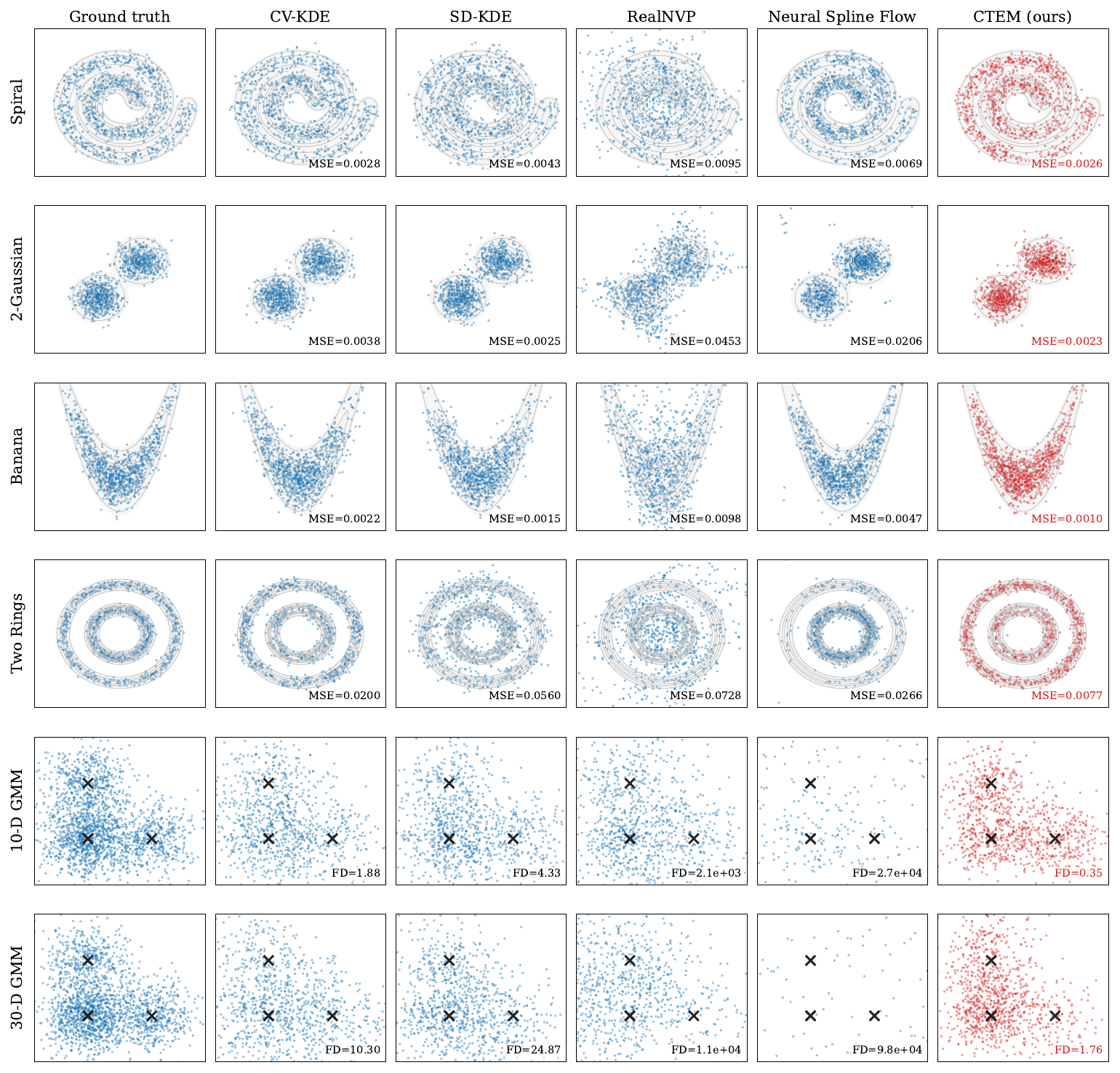}
\caption{Score-driven Langevin samples. Top rows: 2-D benchmarks with
ground-truth density contours. Bottom rows: 10-D and 30-D GMM samples projected
onto $(x_1,x_2)$.}
\label{fig:continuous-samples}
\end{figure}

Figure~\ref{fig:continuous-samples} shows that CTEM produces samples that align
well with the target support and modes. Kernel methods tend to smooth across
disconnected components or under-cover modes, while flow baselines produce
diffuse samples or artifacts induced by inaccurate scores. These sampling
results are consistent with the density MSE and Fisher-divergence comparisons
in Table~\ref{tab:continuous}.

\section{Discrete-variable experimental setup}
\label{app:exp-disc-setup}

This appendix provides implementation details for the discrete experiments in
Section~\ref{subsec:exp-discrete}. The experiments cover two settings:
quantized 2-D density recovery and binary MNIST generation.

\paragraph{Quantized 2-D datasets.}
We evaluate on moons, swissroll, and 8-Gaussians. Moons are generated by
\texttt{sklearn.make\_moons} with noise $0.05$. Swissroll is generated by
\texttt{sklearn.make\_swiss\_roll} with noise $0.5$ and projected to the
$(x,z)$ plane. The 8-Gaussians distribution is an equal-weight mixture of eight
Gaussians arranged on a circle of radius $2$, with component standard deviation
$0.1$. Each dataset is quantized onto a $91{\times}91$ grid, yielding
$8281$ discrete states. We draw $n=20000$ samples per dataset.

\paragraph{Toy-model parameterization and optimization.}
For all 2-D toy experiments,  the comparison graph is the
4-neighbor grid. All methods are trained with full-batch Adam for $100$k iterations using learning rate $5{\times}10^{-4}$.

\paragraph{Binary MNIST.}
We binarize MNIST to obtain a $28{\times}28$ binary state space with alphabet
size $2$ and dimension $784$. The model is a noise-conditional U-Net that takes
a noisy binary image $z_\sigma$ and noise level $\sigma$ as input and outputs a
scalar energy. Local energy differences are computed by evaluating the energy
on single-bit flips. The noise schedule contains $13$ log-spaced noise levels
in $[0.005,1.5]$, which determine the per-pixel Bernoulli flip probability. We train for $300$ epochs with batch size $256$, Adam learning rate
$5{\times}10^{-5}$, and gradient clipping at $100$. Samples are generated by
reverse-time Metropolis--Hastings from the learned noise-conditional energy. We
draw $1024$ generated samples for evaluation.

We evaluate sample recognizability using an off-the-shelf MNIST classifier.
The reported classifier top-1 score is the mean maximum softmax probability on
generated samples. Real binary MNIST obtains a top-1 score of
$0.992\,[0.989,0.994]$, while CTEM obtains
$0.929\,[0.920,0.937]$, where brackets denote $95\%$ bootstrap confidence
intervals over $1024$ samples.

\section{Mixed continuous--discrete experimental setup}
\label{app:exp-mixed-setup}

\paragraph{Data.}
We consider the mixed product space
$\mathcal{Z}=\mathbb{R}^2\times\{0,\ldots,K-1\}$ and define the ground-truth
joint density as
\begin{equation}
    p(x,y)=\pi_y\,p(x\mid y),
    \qquad
    p(x\mid y)=\mathcal{N}(x;\mu_y,\sigma^2 I_2).
    \label{eq:app-mixed-density}
\end{equation}
In all experiments, we set $K=16$ and $\sigma=1$. The component means are
placed uniformly on a ring,
\begin{equation}
    \mu_y
    =
    \bigl(
        R\cos(2\pi y/K),
        R\sin(2\pi y/K)
    \bigr),
    \qquad
    R=s\sigma,
    \label{eq:app-mixed-means}
\end{equation}
where $s$ controls the centre-to-noise separation; the main experiment uses
$s=2$. The label prior is mildly non-uniform: we draw
$\tilde \pi_y\sim \mathrm{Unif}[1,1.5]$ independently for
$y=0,\ldots,K-1$, and then normalize,
\begin{equation}
    \pi_y
    =
    \frac{\tilde \pi_y}{\sum_{j=0}^{K-1}\tilde \pi_j}.
    \label{eq:app-mixed-prior}
\end{equation}
For the main visualization in Figure~\ref{fig:mixed-joint}, we draw training
samples from \eqref{eq:app-mixed-density} and evaluate the learned model on a
held-out set.

\paragraph{Model and training.}
We parameterize the energy by an MLP
$f_\theta(x,y):\mathbb{R}^2\times\{0,\ldots,K-1\}\to\mathbb{R}$, whose input is
the concatenation of $x$ and a one-hot encoding of $y$. To train CTEM, we use
a joint comparison kernel that perturbs both coordinates simultaneously. Given
a sample $(x,y)$, we draw a random unit vector $u\in\mathbb{S}^1$ and a random
nonzero label shift $\delta\in\{1,\ldots,K-1\}$, and define the comparison
state
\begin{equation}
    \tilde x = x-2\varepsilon u,
    \qquad
    \tilde y = (y+\delta)\bmod K.
    \label{eq:app-mixed-joint-kernel}
\end{equation}
The training loss is then
\begin{equation}
    \widehat{\mathcal{L}}_{\mathrm{mixed}}(\theta)
    =
    \frac{1}{B}
    \sum_{b=1}^B
    \left[
        \tanh\!\left(
            \frac{
                f_\theta(x_b,y_b)-f_\theta(\tilde x_b,\tilde y_b)
            }{2}
        \right)
        -1
    \right]^2.
    \label{eq:app-mixed-loss}
\end{equation}
For the figure in the main text, we train an Energy-CTEM model with this joint
kernel using Adam and standard mini-batch optimization.

\paragraph{Evaluation.}
To visualize the learned mixed distribution, we compare the ground-truth
continuous marginal
\[
    p(x)=\sum_{y=0}^{K-1} p(x,y)
\]
with the learned marginal
\[
    \hat p_\theta(x)\propto \sum_{y=0}^{K-1} e^{f_\theta(x,y)}.
\]
We also inspect the per-label components $\{p(x,y)\}_{y=0}^{K-1}$ and their
learned counterparts, which show whether the model correctly captures the
continuous conditional structure associated with each discrete label.

\begin{figure}[t]
\centering
\includegraphics[width=0.48\linewidth]{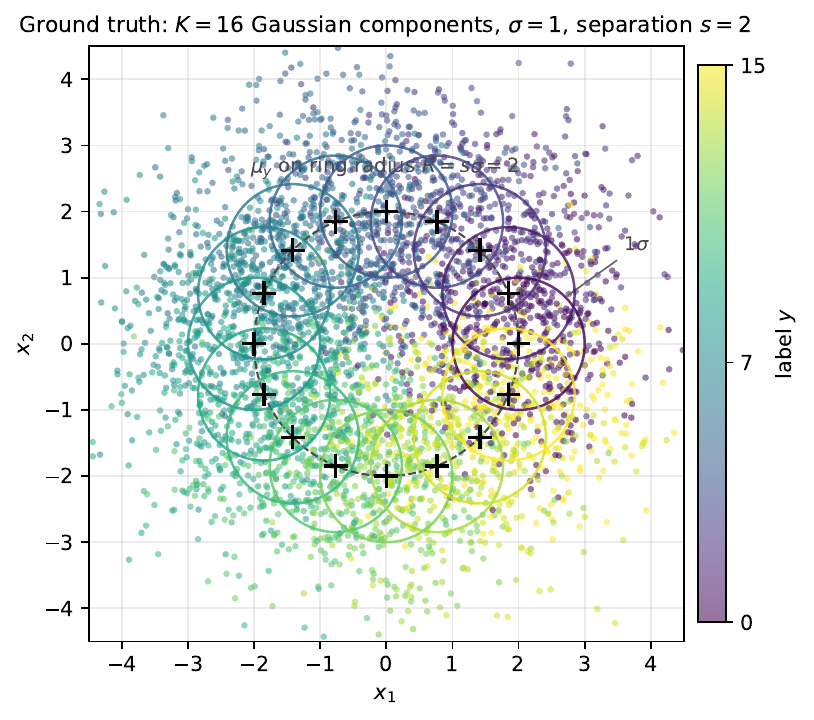}
\caption{Ground-truth mixed distribution on
$\mathbb{R}^2\times\{0,\ldots,15\}$. Each discrete state $y$ corresponds to a
Gaussian component in the continuous variable $x$, with component means placed
uniformly on a ring. Colours indicate the discrete labels, and crosses mark the
component means.}
\label{fig:mixed-gt}
\end{figure}

\end{document}